\documentclass[lettersize,journal]{IEEEtran}
\usepackage{amsmath,amsfonts}
\usepackage{algorithmic}
\usepackage{algorithm}
\usepackage{array}
\usepackage[caption=false,font=footnotesize,labelfont=rm,textfont=rm]{subfig}
\usepackage{textcomp}
\usepackage{stfloats}
\usepackage{url}
\usepackage{verbatim}
\usepackage{graphicx}
\usepackage{cite}
\usepackage{booktabs}
\usepackage{multirow}
\usepackage{threeparttable}
\usepackage[backref]{hyperref}
\begin{document}

\title{Investigating Shift Equivalence of Convolutional Neural Networks in Industrial Defect Segmentation}
\author{Zhen Qu, Xian Tao, \IEEEmembership{Senior Member,~IEEE}, Fei Shen, Zhengtao Zhang, Tao Li
        % <-this % stops a space
\thanks{Zhen Qu and Fei Shen are with the Institute of Automation, Chinese Academy of Sciences, Beijing 100190, China, and also with the School of Artificial Intelligence, University of Chinese Academy of Sciences, Beijing 100049, China.  \par 
Xian Tao and Zhengtao Zhang are with the Institute of Automation, Chinese Academy of Sciences, Beijing 100190, China, also with the Binzhou Institute of Technology, Binzhou, Shandong 256606, China, also with the CAS Engineering Laboratory for Intelligent Industrial Vision, Beijing 100190, China, and also with the School of Artificial Intelligence, University of Chinese Academy of Sciences, Beijing 100049, China (Corresponding e-mail: taoxian2013@ia.ac.cn). \par 
Li Tao is with the Institute of Automation, Gansu Academy of Sciences, Lanzhou 730000, China.
}
\thanks{This work was supported by the National Natural Science Foundation of China under Grant (U21A20482 and 62066004), Beijing Municipal Natural Science Foundation (China) 4212044, the Applied Research and Development Project of Gansu Academy of Sciences (2018JK-01) and project funding from Binzhou Institute of Technology (GYY-ZNJS-2022-ZY-002-2-2022-002).}% <-this % stops a space
}

\markboth{IEEE TRANSACTIONS ON INSTRUMENTATION AND MEASUREMENT}%
{Shell \MakeLowercase{\textit{et al.}}: A Sample Article Using IEEEtran.cls for IEEE Journals}
\maketitle
\begin{abstract}
In industrial defect segmentation tasks, while pixel accuracy and Intersection over Union (IoU) are commonly employed metrics to assess segmentation performance, the output consistency (also referred to equivalence) of the model is often overlooked. Even a small shift in the input image can yield  significant fluctuations in the segmentation results. Existing methodologies primarily focus on data augmentation or anti-aliasing to enhance the network's robustness against translational transformations, but their shift equivalence performs poorly on the test set or is susceptible to nonlinear activation functions. Additionally, the variations in boundaries resulting from the translation of input images are consistently disregarded, thus imposing further limitations on the shift equivalence. In response to this particular challenge, a novel pair of down/upsampling layers called component attention polyphase sampling (CAPS) is proposed as a replacement for the conventional sampling layers in CNNs. To mitigate the effect of image boundary variations on the equivalence, an adaptive windowing module is designed in CAPS to adaptively filter out the border pixels of the image. Furthermore, a component attention module is proposed to fuse all downsampled features to improve the segmentation performance. The experimental results on the micro surface defect (MSD) dataset and four real-world industrial defect datasets demonstrate that the proposed method exhibits higher equivalence and segmentation performance compared to other state-of-the-art methods. Our code will be available at \url{https://github.com/xiaozhen228/CAPS}.
\end{abstract}

\begin{IEEEkeywords}
shift equivalence, industrial defect segmentation, U-Net,  convolutional neural network (CNN),  deep learning.
\end{IEEEkeywords}

\section{Introduction}
\IEEEPARstart{V}{isual} inspection methods based on convolutional neural networks (CNNs) have attracted considerable interest in recent years for industrial quality control of diverse scenes, such as steel surfaces \cite{steel_surfaces}, printed circuit boards \cite{PCB}, rail surfaces \cite{rail_surface}, textured fabrics \cite{fabrics}, and many others. Concurrently, segmentation-based networks for defect detection have become popular due to their ability to provide precise location and contour information of defects \cite{five,six}.
\par 
\begin{figure}[!t]
\centering
\includegraphics[width=0.9\columnwidth]{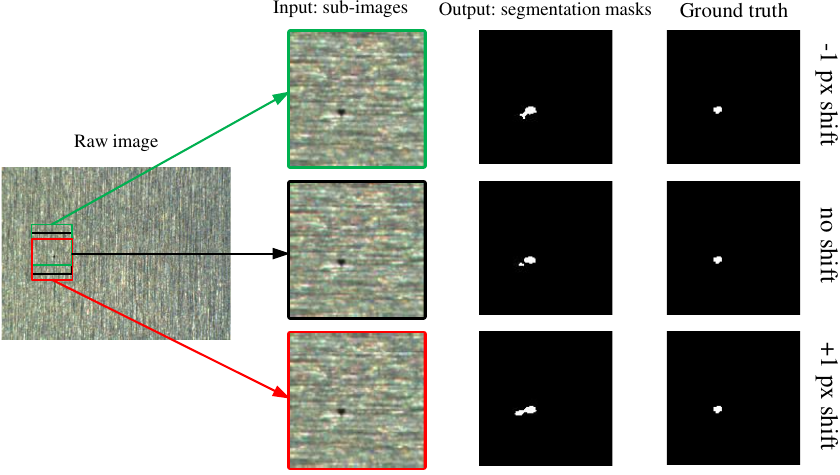}   %pic1.pdf
\caption{Influence of input image translation on output segmentation masks. Initially, the black window in the original image is translated upwards and downwards by one pixel, resulting in the green and red windows, respectively. Subsequently, the images within the three windows are cropped and individually fed into the image segmentation network. The ground truth indicates a defect area of 44 pixels, while the predicted defect areas for the green, black, and red sub-images in the network output are 55, 49, and 63 pixels, respectively.}
\label{fig1}
\end{figure}
However, most segmentation-based networks in defect detection primarily focus on improving segmentation metrics such as pixel accuracy and Intersection over Union (IoU), while neglecting the crucial aspect of output consistency. Output consistency refers to the concept of shift equivalence, which implies that if input images are shifted by a certain number of pixels, the corresponding segmentation masks produced by the network should also exhibit the same pixel offsets. Despite the long-held belief that CNNs inherently possess shift equivalence \cite{seven,eight}, several studies \cite{BlurPool,APS,LPS} have revealed that input translation significantly affects the segmentation outcomes, especially in the industrial inspection field. To shed light on the issue of shift equivalence in CNNs, Fig. \ref{fig1} visually portrays the impact of input translations on the segmentation masks. The defective raw image is partitioned into three sub-images: green, black, and red, with each pair of adjacent sub-images differing by only one pixel in position. The sub-images are subsequently fed into the segmentation network, yet the resulting segmentation masks exhibit significant disparities. This situation often occurs in the following industrial settings: 1) when the same part is repeatedly captured by machine vision equipment with slight pixel translations due to mechanical deviations, leading to significant fluctuations in segmentation outcomes; 2) in a defective image, the same defects may vary by just a few pixels in position from image to image due to sampling, resulting in highly inconsistent segmentation outcomes. Therefore, the issue of shift equivalence has gained widespread attention among scholars in recent years.

\begin{figure*}[!tb]
	\centering
	\subfloat[]{\includegraphics[width=0.48\columnwidth]{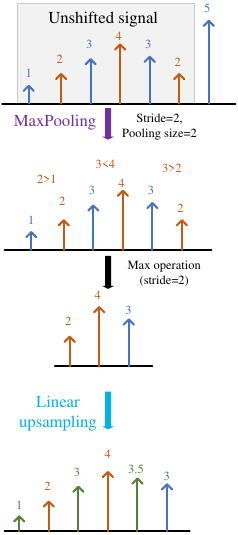}%
		\label{fig2_a}}
	\hfil
	\subfloat[]{\includegraphics[width=0.48\columnwidth]{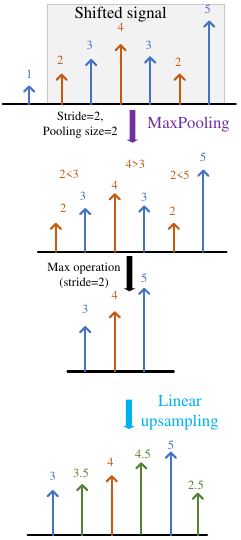}%
		\label{fig2_b}}
	\hfil
	\subfloat[]{\includegraphics[width=0.5\columnwidth]{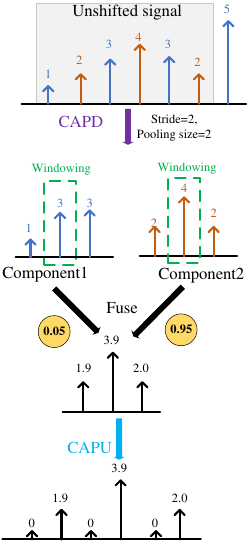}%
	\label{fig2_c}}
	\hfil
	\subfloat[]{\includegraphics[width=0.48\columnwidth]{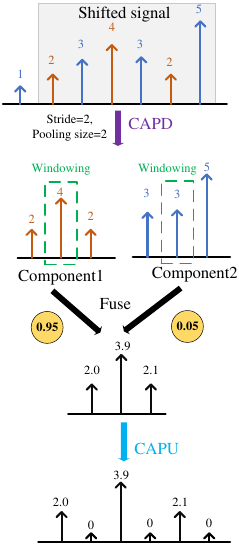}%
	\label{fig2_d}}		
	\caption{A visual comparison of two downsampling methods and their corresponding upsampling techniques based on a one-dimensional signal. (a)  MaxPooling process. (b)  same input signal in Fig. 2(a) after an one-unit leftward translation with the MaxPooling process. (c) the proposed method.  (d)  same input signal in Fig. 2(c) after an one-unit leftward translation with the CAPD process. First, assume that the unshifted signal input to the downsampling layer is [1, 2, 3, 4, 3, 2]. Then, the shifted signal,  when shifted one unit to the left, is represented as [2, 3, 4, 3, 2, 5]. The stride and pooling size during downsampling are both set to 2. As depicted in Figs. 2(a) and (b), the downsampling results after MaxPooling for the unshifted signal and the shifted signal are [2, 4, 3] and [3, 4, 5], respectively. It is observed that the results of MaxPooling are quite different ([2, 4, 3] vs. [3, 4, 5]).  However, the proposed CAPD keeps the results similar after downsampling ([1.9, 3.9, 2.0] vs. [2.0, 3.9, 2.1] ) as shown in Figs. 2(c) and (d). Specifically, the input signal is first sampled into two components, Component 1 and Component 2, according to the parity index. Then, the boundary elements of the components are filtered by adding a window and the corresponding weights are acquired from the component attention module. Lastly, the different components are weighted and fused to obtain the downsampled results.  }
	\label{fig2}
\end{figure*}
\par 
The strategies to address the problem of shift equivalence in CNNs can be broadly categorized into learning-based and design-based approaches \cite{chen}. The former primarily focuses on enhancing network robustness through a data-driven approach, such as data augmentation. However, its segmentation performance shows a significant decline in the test set. The latter strategy seeks to redesign the network architecture in order to rectify the lack of equivalence in CNNs without relying on data. One key factor contributing to the loss of shift equivalence in CNNs is the downsampling layers, such as pooling layers and strided convolution layers, which violate the Nyquist-Shannon sampling theorem as highlighted by Zhang \cite{BlurPool}. Therefore, it is meaningful to devise a new downsampling technique to cover traditional sampling layers such as MaxPooling, ensuring that the downsampled feature maps remain as similar as possible before and after image translation. Currently, two new downsampling design techniques have been proposed to reduce the disparities in downsampled feature maps, namely, anti-aliasing and component selection. Anti-aliasing-based methods, exemplified by BlurPool \cite{BlurPool}, aim to minimize differences between adjacent pixels by incorporating a low-pass filter (LPF) to remove high-frequency components from the image. However, these methods face limitations in nonlinear systems, especially when nonlinear activation functions like ReLU are present in the network \cite{azulay}. On the other hand, the component-selection-based methods, represented by adaptive polyphase sampling (APS) \cite{APS} and learnable polyphase sampling (LPS) \cite{LPS}, were designed to select the same components as much as possible during downsampling before and after translation, thereby achieving structural equivalence in CNNs. Although the component-selection-based methods have demonstrated effectiveness in improving shift equivalence, they have not taken into account the variations in image boundaries that occur when input images are shifted in the manner depicted in Fig. \ref{fig1}. These variations result in random pixels at the image boundaries, making it challenging to ensure the similarity of downsampled results or the selection of identical components before and after image translation, thereby further constraining the shift equivalence. Furthermore, selecting specific component implies discarding the remains, which has an impact on the segmentation performance.  
\par
To address this issue, a novel method called component attention polyphase sampling (CAPS) is proposed in this paper. CAPS contains two essential layers, namely component attention polyphase downsampling (CAPD) and component attention polyphase upsampling (CAPU), to replace the conventional downsampling and upsampling layers in CNNs. The CAPD aims to ensure maximum similarity of the downsampled results before and after image translation. It mainly consists of three parts: a polyphase downsampling process, an adaptive windowing (AW) module, and a component attention (CA) module. Initially, the input image undergoes polyphase downsampling, generating four components with half of the original spatial resolution. These components are then extracted as features and sequentially processed through the AW and CA modules to generate attention weights corresponding to each component. The downsampled results are finally achieved by fusing the different initial component features with the attention weights. The AW module effectively mitigates the boundary effect caused by shifts in images, thereby enhancing the consistency of downsampled features. On the other hand, the CA module captures global features of the components through global average pooling (GAP) and employs one-dimensional convolution to facilitate component-wise attention, leading to significant improvement in defect segmentation performance through the fusion of all downsampled components. Corresponding to the implementation of downsampling using CAPD, CAPU restores the downsampled features to their original spatial positions, thereby ensuring shift equivalence in segmentation networks. 
\par 
Fig. \ref{fig2} provides a visual comparison of two downsampling methods and their corresponding upsampling techniques based on a one-dimensional signal. It can be seen from Fig. \ref{fig2} that MaxPooling selects the maximum value at fixed positions as the downsampled result. When the input undergoes translation, the maximum value within the corresponding pooling region has already changed, leading to significant alterations in the downsampled result. However, the proposed CAPS samples the input signal into two components based on its odd and even indices. When the input undergoes translation, only the odd and even indices are swapped, and the values within each component remain the same. The fusion results of CAPD are also largely identical for the similarity of components.
\par
Our contributions are summarized as follows: 
\par
1. A pair of down/upsampling layers called CAPS is proposed to address the shift equivalence problem and can serve as alternatives for conventional downsampling and upsampling layers in CNNs. To the best of our knowledge, this work is the first to investigate the issue of shift equivalence in the field of industrial defect segmentation, considering the boundary variations caused by image translation and leveraging all downsampled component features. 
\par 
2. CAPD, a novel downsampling layer, is designed to maximize the similarity of downsampled results before and after image translation. The AW module mitigates boundary effect, while the CA module integrates different components to enhance segmentation performance.
\par 
3. The proposed method outperforms other state-of-the-art anti-aliasing-based and component-selection-based methods in both shift equivalence and segmentation performance on the micro surface defect (MSD) dataset and four real-world industrial defect datasets.
\par 
\section{RELATED WORK}
In this section, defect detection relying on segmentation networks is first reviewed. Subsequently, a comprehensive works related to shift equivalence is introduced.
\subsection{Defect Segmentation}
Defect segmentation, a crucial technique in defect detection, has gained significant traction in real-world industrial vision scenarios. Currently, deep learning methods (e.g. FCN \cite{FCN}, DeepLabv3+ \cite{deeplabv3+}, U-Net \cite{unet}, etc.) have emerged as popular choices for industrial vision defect segmentation due to their robust characterization and generalization capabilities. Wang et al. \cite{Wang} employed a three-stage FCN to enhance the accuracy and generalization ability of defect segmentation in tire X-ray images. To address the limited sample issue in print defect datasets, Valente et al. \cite{Valente} utilized computer graphics techniques to synthesize new training samples at the pixel level, resulting in commendable segmentation performance using DeepLabv3+. Miao et al. \cite{Miao} designed a loss function of U-Net network based on the Matthews correlation coefficient to tackle the challenges posed by limited data volume and imbalanced sample categories. To alleviate the contextual feature loss caused by multiple convolutions and pooling during the encoding process, Yang et al. \cite{Yang} introduced a multi-scale feature fusion module into the U-Net network. They integrated a module named bidirectional convolutional long short-term memory block attention into the skip connection structure, effectively capturing global and long-term features. Zheng et al. \cite{Zheng} devised a novel Residual U-Structure embedded within U-Net, complemented by a coordinate attention module to integrate multi-scale and multi-level features. In summary, current methodologies predominantly focus on extracting rich features to enhance defect segmentation performance, often overlooking the significance of shift equivalence. Therefore, it holds great significance to study shift equivalence in CNNs.

\begin{table*}[!t]
  \caption{REPRESENTATIVE WORK FOR IMAGE SHIFT EQUIVALENCE}
  \centering
  \label{tab1}
  \renewcommand{\arraystretch}{1}
  \begin{tabular}{c>{\centering\arraybackslash}m{2.4cm}>{\centering\arraybackslash}m{2.4cm}>{\centering\arraybackslash}m{3cm}>{\centering\arraybackslash}m{4cm}}
    \toprule
    Method   & Venue        & Whether to consider image boundaries & Key point           & Major shortcoming                                                              \\ \midrule
    Data Augmentation & ——   & No                                & Expending the training dataset       & Cannot generalize to samples that do not appear in the training set                                    \\
   \midrule
   DUNet \cite{DUNet}  &  Knowledge-Based Systems: 2019  &  No   &   Deformable convolution      &  Time-consuming and data-dependent    \\   
   \midrule 
    BlurPool \cite{BlurPool} & PMLR:2019   & No                                & Anti-Aliasing       & Sensitive to nonlinear activation functions                                    \\
   \midrule
   PBP \cite{PBP}    &  IEEE T ULTRASON FER: 2022      &    No             &  Anti-Aliasing          &  Sensitive to nonlinear activation functions    \\
   \midrule
    APS \cite{APS}      & CVPR:2021    & No                                & Component Selection & Vulnerable to image boundaries; Cannot select the downsampled component adaptivey \\
   \midrule
    LPS \cite{LPS}      & NeurIPS:2022 & No                                & Component Selection & Vulnerable to image boundaries; Cannot utilize the full downsampled component \\
    \midrule
    MWCNN \cite{MWCNN}    &       CVPR:2018    &   No         &   Discrete wavelet transform and inverse wavelet transform  &   Neglecting the order of concatenating downsampled components  \\
    \midrule
    Ours (CAPS)     &    ——          & Yes                               & Component Fusion    &      ——                                                                          \\ \bottomrule
  \end{tabular}
\end{table*}
\subsection{Shift Equivalence}
In contrast to shift equivalence in segmentation tasks, shift equivalence in image classification - also known as shift invariance - has been extensively studied with some promising results \cite{Simoncelli,Zou}. The strict distinction between shift equivalence and shift invariance is presented in Section \ref{PROBLEM DESCRIPTION}. Learning-based approaches such as data augmentation \cite{augmentation} and deformable convolution \cite{deformable,DUNet} are data-driven ways to improve shift equivalence. Deformable convolution introduces additional learnable offset parameters within standard convolutional operations, which can acquire adaptive receptive fields and learn geometric transformations automatically.  Deformable U-Net (DUNet) \cite{deformable} is a typical application of deformable convolution in the field of image segmentation. It aims to replace  the traditional convolutional layers in U-Net \cite{unet} with a deformable convolutional block to make the network adaptive in adjusting the receptive field and sampling locations according to the segmentation targets. However, this approach is time-consuming and relies heavily on training data, making it difficult to apply to real industrial scenarios. Modern CNNs lack equivalence unless the input image is shifted by an integer multiple of the downsampling factor, which is impractical \cite{Simoncelli}. As a result, the redesign of traditional downsampling methods has emerged as an effective strategy for improving equivalence.
\par  
Anti-aliasing has proven to be an effective approach for enhancing equivalence by addressing the violation of Nyquist-Shannon sampling theorem during downsampling \cite{BlurPool}. In this approach, the MaxPooling layer (stride 2) was partitioned into a dense MaxPooling layer (stride 1) and a naive downsampling layer (stride 2). Additionally, a low-pass filter (LPF) was utilized to blur the features after dense maxpooling layer, effectively mitigating aliasing effect. Zhang's work \cite{BlurPool} paved the way for further advancements, such as the utilization of adaptive low-pass filter kernels based on BlurPool \cite{Zou}, and the design of Pyramidal BlurPooling (PBP) structure to gradually reduce the size of the LPF kernel at each downsampling layer \cite{PBP}. Despite anti-aliasing-based approaches offering some degree of improvement in shift equivalence, their ability to address the equivalence problem remains limited. First, aliasing can only be completely eliminated in linear systems by blurring prior to downsampling, which contradicts the prevalent use of nonlinear activation functions (e.g., ReLU) in current CNNs \cite{azulay}. Second, it employs LPF before downsampling, resulting in a trade-off between image quality and shift equivalence \cite{Lenc}.  
\par 
Another elegant approach for improving downsampling is component-selection-based methods. These methods involve downsampling the input using fixed strides to obtain a set of downsampled components, followed by a strategy to select one of the components as the downsampled feature map. By ensuring consistent selection of the same component for each downsampling operation, shift equivalence in CNNs can be achieved by restoring the feature map to its corresponding position during upsampling. Chaman et al. \cite{APS} employed the max-norm polyphase component during downsampling in their APS method, proving complete shift equivalence under circular shifts. Gomez et al. \cite{LPS} utilized a small neural network to adaptively select components, thereby further improving segmentation performance without compromising equivalence in LPS. Both APS and LPS utilized LPF before downsampling and after
upsampling, as ref. \cite{APS} indicated that anti-aliasing can further improve the segmentation performance significantly. However, APS and LPS did not achieve the expected performance in terms of equivalence when faced with common shifts in input images. This can be attributed to two main reasons. First, boundary variations resulting from image translation are not considered during the downsampling process, making it challenging to ensure consistent component selection before and after image translation. Second, selecting a single component as the downsampled result discards the majority of features, leading to reduced segmentation performance. In order to solve the problem of information loss during downsampling, Liu et al. \cite{MWCNN} proposed a novel multi-level wavelet CNN (MWCNN) method, which employs discrete wavelet transform (DWT) and  inverse wavelet transform (IWT) to downsample and upsample the features respectively. However, MWCNN concatenates the four components directly after DWT while ignoring their order, resulting in the loss of shift equivalence. 
\subsection{Positioning of Our CAPS}
The proposed approach in this paper focuses on a structural redesign of the network to enhance shift equivalence in the segmentation task. Specifically, CAPS improves the equivalence of the segmentation networks by redesigning the downsampling layer CAPD and the upsampling layer CAPU. Table \ref{tab1} analyzes representative methods and ours in terms of whether to consider image boundaries, key points, and major shortcomings. Compared to other methods, the proposed method considers the variations in image boundaries due to translation, thereby enhancing the consistency of downsampled results before and after translation. Moreover, unlike other methods that rely on selecting a single component after downsampling, CAPS incorporates the fusion of multiple downsampled component features, thereby improving the overall segmentation performance.
\section{PROBLEM DESCRIPTION}
Here, we provide definitions of shift equivalence and shift invariance, along with graphical examples to illustrate the issues first. Then, the boundary problem when input translation occurs is pointed out and a preliminary solution is proposed. To enhance readability, one-dimensional signals are employed to illustrate this section. It is worth noting that a two-dimensional image can be seen as an extension of a one-dimensional signal. 
\begin{figure}[H]
\centering
\includegraphics[width=0.95\columnwidth]{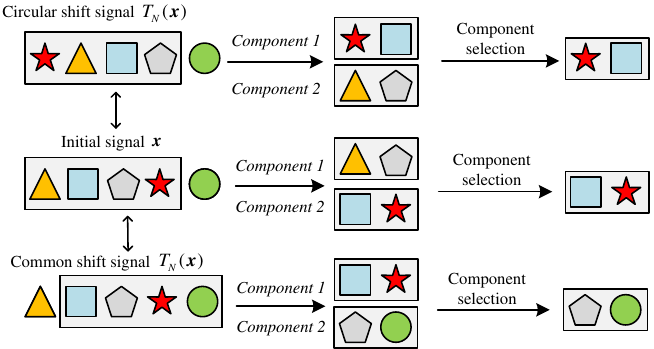}
\caption{The downsampling method based on component selection for shift equivalence. The rectangular region in the middle row represents the initial signal. \textit{Component 1} and \textit{Component 2} sample the odd and even positions of the input signal, respectively. The component-selection-based approaches select one of the two components as the result of downsampling according to a specific strategy.}
\label{fig3}
\end{figure}
\subsection{Definitions of shift invariance and shift equivalence}
\label{PROBLEM DESCRIPTION}
To clarify the concept of shift equivalence, it is important to first distinguish it from the related concept of shift invariance. Shift invariance refers to a mapping that remains constant before and after the input is shifted, and is commonly used to indicate that the translation of an input image does not affect the final predicted class in image classification tasks. For an input signal $\boldsymbol{x}$ and its shifted vision $T_N(\boldsymbol{x})$, an operation $\tilde{f}$ considered to be shift-invariant can be defined as:
\begin{equation}
\label{eq1}
\tilde{f}(\boldsymbol{x}) = \tilde{f}(T_{N}(\boldsymbol{x}))
\end{equation}
where $N$ denotes the number of signal shifts in the circular or common shift way. Shift equivalence, however, dictates that the output should shift concurrently with the input, which is commonly utilized to describe image segmentation tasks. Accordingly, an operation $\tilde{f}$ is shift-equivalent can be expressed as:
\begin{equation}
\label{eq2}
T_{N}(\tilde{f}(\boldsymbol{x})) = \tilde{f}(T_{N}(\boldsymbol{x}))
\end{equation}   
\par 
\subsection{Description of shift equivalence and boundary effect}
Fig. \ref{fig3} illustrates the downsampling methods based on component selection for shift equivalence, such as APS and LPS. In the second row, the initial signal $\boldsymbol{x}$ comprises four elements: an orange triangle, a blue square, a grey pentagon, and a red pentagram. The signal $\boldsymbol{x}$ is then sampled into \textit{component 1} and \textit{component 2} according to the odd/even indices, one of which is eventually selected as the result of downsampling. The components acquired from initial signal $\boldsymbol{x}$  and its circular shift vision $T_{N}(\boldsymbol{x})$ (first row) contain the same elements, only in a different order. Therefore, it can theoretically guarantee shift equivalence if the same components are selected during downsampling when the input images are shifted, as proved by ref. \cite{LPS}. Nevertheless, as depicted in the third row of Fig. \ref{fig3}, in the case of a common shift, its \textit{component 1} corresponds precisely to \textit{component 2} of the initial signal, while its \textit{component 2} manifests an additional element (represented by the green circle) absent in the initial signal. Moreover, as shown in the third row, the orange triangle in the initial signal  does not appear in the common shift version $T_{N}(\boldsymbol{x})$. Hence, random variations in the input signal boundaries cause variability in the selection of the downsampled components, resulting in the loss of full shift equivalence. Due to the boundary variations that occur during common shifts, Eq. \ref{eq2} considers only the unchanged part of the input image before and after the translation. It is important to note that, unless explicitly specified, all shifts of input images in this paper specifically pertain to common shift rather than circular shift. 
\par 
As shown in Fig. \ref{fig3}, the previous component-selection-based method only kept a certain component as the result of downsampling, which does not make full use of all components. So fusing all components in a set of specific weights is a good way to exploit full characteristics. Moreover, for the boundary variations that make downsampled results uncertain, an effective way to reduce the variation of image boundaries is to adaptively crop feature boundaries based on the input dimension.
\section{PROPOSED METHOD}
 In the following text, the pipeline of the proposed method is introduced and then the design details of CAPD and CAPU are expressed. Following that, the equivalence proof regarding CAPS is provided. Lastly, the loss function is specified.
\subsection{Pipeline}
The U-Net is widely used in industrial defect segmentation for its strong segmentation capabilities and simple architecture \cite{Miao,Yang,Zheng}. It not only has fast inference speed to meet the demand of industrial real-time segmentation, but also has a very good guarantee of segmentation performance for its skip connection structure to fuse more level features. Moreover, the symmetric downsampling and upsampling structure of U-Net can be easily replaced with the proposed CAPS to verifying its performance. Thus, the U-Net is adopted as the base model in this paper and other compared methods have been further improved on this basis to enhance shift equivalence. 
\par
\begin{figure}[H]
\centering
\includegraphics[width=0.9\columnwidth]{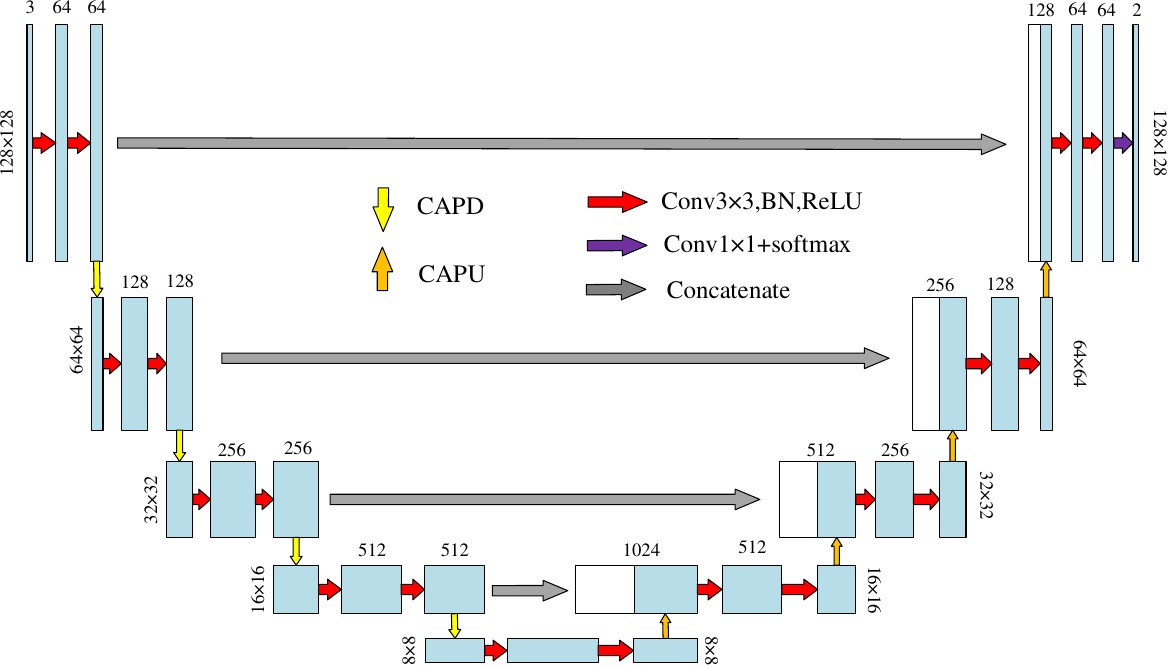}   % pic3.pdf
\caption{The network architecture of our proposed method. Compared to the standard U-Net network, only the downsampling and upsampling layers were replaced with CAPD (yellow arrows) and CAPU (orange arrows), respectively.}
\label{fig4}
\end{figure}

 
The standard U-Net consists of an encoder for feature extraction and a decoder for recovering the original spatial resolution. By using a \textit{Crop} and a \textit{Skip connection} operations, lower-level features are concatenated with higher-level features along the channel dimension to fuse more informative features. CAPS is incorporated into the network architecture as illustrated in Fig. \ref{fig4}. Unlike the standard U-Net, the CAPD layer is designed to perform downsampling instead of MaxPooling in the encoder. Similarly, the CAPU layer replaces the transposed convolution for upsampling the features in the decoder.
\par 
Let us denote $\mathbf{X}\in \mathbb{R}^{H\times W \times C} $ as an input defect image and the output of the model $\hat{\mathbf{Y}}\in \mathbb{R}^{H\times W \times 2}$ can be modeled as:
\begin{equation}
\label{eq3}
\hat{\mathbf{Y}} = f_{model}(\mathbf{X},\theta)
\end{equation}
where $f_{model}:\mathbb{R}^{H\times W \times C}\mapsto \mathbb{R}^{H\times W \times 2}$ , $\theta$ are parameters in the proposed model, and the elements in $\hat{\mathbf{Y}}$ are constrained to binary values of 0 or 1, symbolizing the background and the defect, respectively. Through the process of back-propagation, performed during the training phase, the optimal parameters $\theta^{*}$ of the proposed model can be expressed as below:
\begin{equation}
\label{eq4}  
\theta^{*}  =\underset{\theta}{{\arg\max}} \, l(\hat{\mathbf{Y}},\mathbf{Y})
\end{equation}
in which $\mathbf{Y}$ denotes the ground truth of the input image  $\mathbf{X}$ and $l$ represents the loss function as outlined in Section \ref{Loss Function}.
\subsection{Component Attention Polyphase Downsampling (CAPD) Layer}
\begin{figure*}[!t]
\centering
\includegraphics[width=1.8\columnwidth]{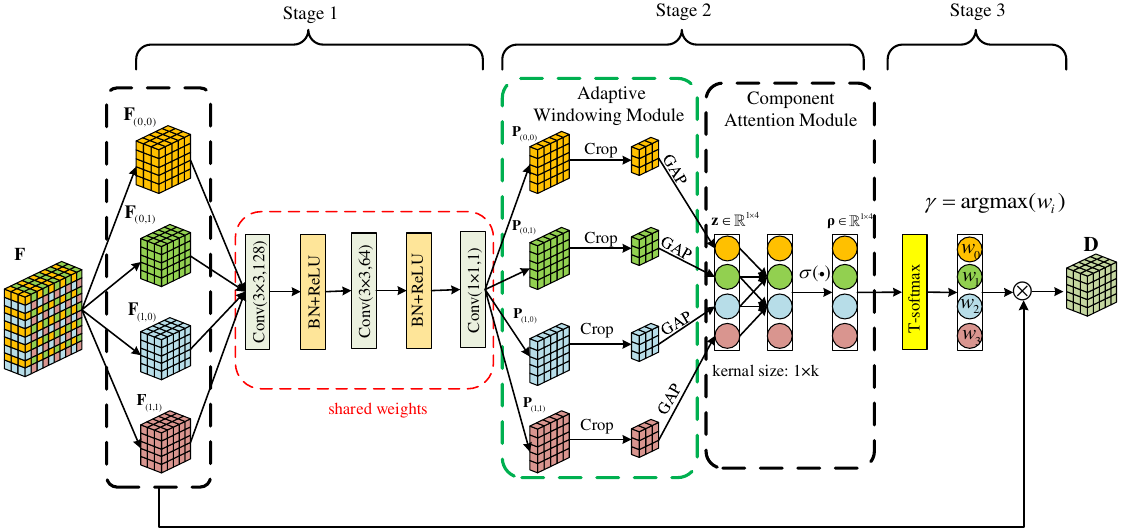}
\caption{The diagram of our proposed CAPD layer.  In the first stage, the input features are first polyphasically sampled into four components according to odd and even indices. Then, these components are fed into a neural network with shared weights for feature extraction. The extracted features are used in the second stage to generate initial component weights that characterize the different levels of importance through the adaptive windowing and component attention modules, respectively. In the third stage, the initial weights are processed by \textit{T-softmax} function to obtain the final weights and the different components are weighted and fused using the final weights to acquire the final downsampled features.}
\label{fig5}
\end{figure*}

The architecture of CAPD is visualized in Fig. \ref{fig5} and the downsampling process is divided into three stages. The first stage is a polyphase downsampling process and the output four downsampled components are fed into a small neural network for feature extraction. Moving on to the second stage, the feature maps derived from the first stage are processed through the AW module and CA module. This stage is to adaptively remove uncertain feature boundaries and determine the initial weights for the different components. The final downsampled result is acquired in the third stage by weighting and fusing the initial four components.

\textbf{Polyphase downsampling and feature extraction.}
First, the input features undergo polyphase downsampling, resulting in a reduction of the original spatial resolution by half.  They are partitioned into four components based on their spatial locations. Given an input feature  $\mathbf{F}\in \mathbb{R}^{h\times w \times c} $, four components are achieved by polyphase downsampling in the spatial dimension at equal intervals:
\begin{equation}
\label{eq5}  
\mathbf{F}_{(i,j)}[x,y,z] = \mathbf{F}[2x+i,2y+j,z]
\end{equation}
where $\mathbf{F}_{(i,j)}\in \mathbb{R}^{\frac{h}{2}\times \frac{w}{2} \times c},i,j\in \{0,1\} $  denotes the four downsampled components as illustrated in Fig. \ref{fig5}. These components are then passed through two convolutional layers with [3×3, 128] and [3×3, 64] convolutional kernels, respectively. To fully extract their features, a [1×1, 1] convolution kernel is then utilized to compress the features in the channel dimension. The resulting output $\mathbf{P}_{(i,j)}\in \mathbb{R}^{\frac{h}{2}\times \frac{w}{2} \times 1}$  is subsequently employed as the input for the subsequent AW module.
\par
\textbf{AW module and CA module.}
The process of windowing in the AW module is expressed as follows:
\begin{subequations}\label{eq6}
\begin{gather}
\mathbf{z}=Cat\{ GAP\{\mathbf{P}_{(i,j)}[hs:-hs,ws:-ws,:] \}\} \\
hs= \text{int}\left(\frac{h}{2} \times \beta \times \frac{1}{2}\right) \\
ws= \text{int}\left(\frac{w}{2} \times \beta \times \frac{1}{2}\right)
\end{gather}
\end{subequations}
where $\mathbf{z}\in \mathbb{R}^{1\times 4}$  denotes the output of the AW module and $\beta$ corresponds to the proportion of the cropped feature boundaries. The symbols $GAP$ and $Cat$ refer to the operations of global average pooling and concatenation, respectively. After conducting an ablation analysis of hyperparameters, the hyperparameter $\beta$ was set to 0.25 to achieve optimal equivalence.
 \par 
The CA module is intended to enable cross-component interaction for feature fusion motivated by \cite{ECA}. The initial weights $\boldsymbol{\rho}\in \mathbb{R}^{1\times 4}$ of the four components from the CA module can be mathematically expressed as:
\begin{equation}
\label{eq7}  
\boldsymbol{\rho} = \sigma(\mathbf{H}^{(k)} \ast \mathbf{z})
\end{equation}
where $\sigma(\cdot)$ represents the sigmoid function defined as $\sigma(x)=1/(1+e^{-x})$, while $\mathbf{H}$ denotes a one-dimensional convolution kernel with a size of $k$. In this paper, the hyperparameter $k$ was set to 2 since only four components of global features are required to interact and get attention weights. To summarize, the CA module serves two primary purposes: 1) acquiring initial weights for the fusion of components in the third stage through attention mechanisms, and 2) facilitating end-to-end learning of the network by ensuring that the corresponding polyphase downsampled components receive similar initial weights before and after the translation of input images.
\par
\textbf{Fusion of components.}
The utilization of component fusion approaches evidently demonstrates their advantage in improving segmentation performance compared to selecting a single component. However, every coin has two sides. Sometimes when the initial weights of different components are similar, the model fails to concentrate on a specific component. Hence, to enhance the consistency of the downsampled feature maps, a more discriminative component weight is necessary. In this regard, \textit{T-softmax} function \cite{T-Softmax} is incorporated to adjust the weights $\rho_{i}$ resulting in a larger variance. The final component weight $w_i$ is calculated as follows:
\begin{equation}
\label{eq8}  
w_i = \frac{\exp(\rho_{i}/T)}{\sum_{j=0}^{3}\exp(\rho_{j}/T)}, i=0,1,2,3
\end{equation}
where $T$ denotes the temperature coefficient, set $10^{-3}$  according to the ablation experiments to balance the shift equivalence and segmentation performance. Following this, the result of downsampling is denoted as:
\begin{equation}
\label{eq9}  
 \mathbf{D}_{c} = w_0\mathbf{F}_{(0,0)} + w_1\mathbf{F}_{(0,1)} +w_2\mathbf{F}_{(1,0)} +w_3\mathbf{F}_{(1,1)} 
\end{equation}
where $\mathbf{D} \in \mathbb{R}^{\frac{h}{2}\times \frac{w}{2} \times c}$  is the final result of downsampling. In essence, the CAPD is ultimately designed to make the images after translation have similar downsampled feature maps $\mathbf{D}$  as possible without losing feature information. Eventually, these downsampled feature maps are upsampled using the CAPU according to upsampling factor $\gamma$, which keeps track of the positions that require restoration during the upsampling process:
\begin{equation}
\label{eq10}  
\gamma = \arg\max(w_i),i=0,1,2,3
\end{equation}
\subsection{Component Attention Polyphase Upsampling (CAPU) Layer}
\begin{figure}[H]
\centering
\includegraphics[width=\columnwidth]{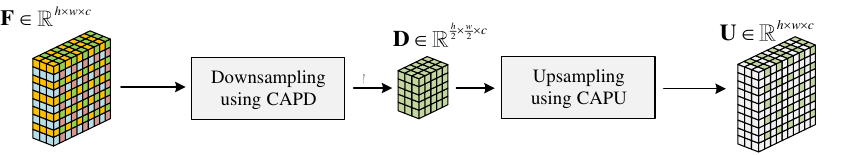}
\caption{One complete downsampling and upsampling process.}
\label{fig6}
\end{figure}

The upsampling process of CSPU is straightforward, involving the placement of the components obtained from downsampling into predetermined spatial positions in the upsampled feature maps. Moreover, the remaining positions in the upsampled feature map are filled with zeros to minimie the uncertainty during the upsampling process. Fig. \ref{fig6} illustrates a complete downsampling and upsampling process, assuming that the input feature is $\mathbf{F}\in \mathbb{R}^{h\times w \times c}$. Specifically, the input $\mathbf{F}$ first undergoes the CAPD layer, yielding the downsampled feature $\mathbf{D} \in \mathbb{R}^{\frac{h}{2}\times
 \frac{w}{2} \times c}$  and the sampling factor $\gamma$  corresponding to the maximum weight. Then the upsampled result 
 $\mathbf{U}\in \mathbb{R}^{h\times w \times c}$ is denoted as:
 
 \begin{equation}
 	\label{eq11}
 {\mathbf{U}_{c}} =  T_{m,n}(U_2(\mathbf{D}_{c}))
 \end{equation}

where $m$ and $n$ map $\gamma$ to a two-dimensional position encoding, which can be achieved through a simple binary encoding process represented by:
\begin{equation}
\label{eq12}  
 mn = \phi(\gamma)
\end{equation}
where $m$ corresponds to the first bit of the encoding result and $n$ represents the second bit of the encoding result. The function $\phi$ converts a decimal number into a binary code. $T_{m,n}(\cdot)$ represents translating the input feature by $m$ and $n$ pixels in the $x$ and $y$ axes, respectively and $U_2$ is the is a conventional upsampling operation. $U_2(\mathbf{D}_c)=\mathbf{Z}[x,y,z]$ can be calculated as:
\begin{equation}
\label{eq13} 
{\mathbf{Z}[x,y,z]} =  \begin{cases}
\mathbf{D}_c[x/2,y/2,z], \text{when x and y are even} \\ 
 0,\quad {\text{otherwise.}} 
\end{cases}
\end{equation}
Following APS and LPS, LPF is also added before CAPD and after CAPU to improve the segmentation performance. Moreover, in the next subsection, we can show that the CAPS is completely equivalent when the boundaries of the features are not considered and $T\to 0$.
\subsection{Proof of shift equivalence for CAPS}
For the simplicity of the proof, the channel dimensions of the input features are not considered and the stride is set to 2. The final result can also be easily generalized to multiple channels and other stride. In addition, the boundaries of the features (i.e. pixels newly entering and moving out of the subimage during the translation process) are not be considered because the feature changes of the boundaries are random and unpredictable when common shift is performed on the image.
Corresponding to $U_2$ in Eq. \ref{eq11}, let $D_2$ represent the traditional downsampling operation and $\mathbf{Q} = D_2(\mathbf{F})$ is given by $\mathbf{Q}[x,y] = \mathbf{F}[2x,2y] $. It is clear that $\mathbf{Q}$ is a two-dimensional simplified version of the first downsampled component $\mathbf{F}_{(0,0)}$ in Eq. \ref{eq5}, and the other downsampled components can be expressed as ${\{\mathbf{F}_{(i,j)}\}_{i,j=0}^{1}}$, where $\mathbf{F}_{i,j} = D_2(T_{-i,-j}(\mathbf{F})) = \mathbf{F}[2x+i,2y+j]$. Let us denote $D_{2}^{c}(\cdot)$ and $U_{2}^{c}(\cdot)$ as the CAPD and CAPU operator, which are defined as:
\begin{equation}
\label{eq14} 
\mathbf{D}_c = \mathbf{F}_{m,n} = D_{2}^{c}(\mathbf{F}) = D_2(T_{-m,-n}(\mathbf{F})) 
\end{equation}
\begin{equation}
	\label{eq15}
 U_{2}^{c}(\mathbf{D}_c,m,n) = T_{m,n}(U_2(\mathbf{D}_c))
\end{equation}
where $m$ and $n$ denotes the index of component with the highest weight as indicated in Eq. \ref{eq12}. Note that the conditional equality of $\mathbf{D}_c = \mathbf{F}_{m,n}$ in Eq. \ref{eq14} is the temperature coefficient $T\to 0$ in Eq. \ref{eq8}. We can now show that $U_{2}^{c} \circ D_{2}^{c}$ is fully equivalent when variations in the image boundaries due to translation are not considered and $T\to 0$: 
\begin{equation}
	\label{eq16}
U_{2}^{c} \circ D_{2}^{c} (\widetilde{\mathbf{F}}) = T_{s_x,s_y}(U_{2}^{c} \circ D_{2}^{c}(\mathbf{F})), \forall s_x,s_y \in \mathbb{Z}
\end{equation}
where $\widetilde{\mathbf{F}} = T_{s_x,s_y}(\mathbf{F})$ represents the result of translating the input $\mathbf{F}$ by $s_x$ and $s_y$ pixels along the x-axis and y-axis directions, respectively.
\par
\textit{Proof}. Let $m,n$ and $\widetilde{m},\widetilde{n}$ denote the component index corresponding to the maximum weight obtained with $\mathbf{F}$ and $\widetilde{\mathbf{F}}$ as CAPS inputs, respectively. Then assume that $s_x$ and $s_y$ are both odd integers:
\begin{subequations}
\begin{gather}
D_2^c(\mathbf{F}) = D_2(T_{-m,-n}(\mathbf{F}))  \\
D_2^c(\widetilde{\mathbf{F}}) = D_2(T_{-\widetilde{m},-\widetilde{n}}(\widetilde{\mathbf{F}}))
\end{gather}
\end{subequations}
Based on the above properties we can get:
\begin{equation}
	\label{eq18}
U_{2}^{c} \circ D_{2}^{c}(\mathbf{F}) = T_{m,n}U_2D_2(T_{-m,-n}(\mathbf{F}))
\end{equation}
Similarly for the input after translation:
\begin{subequations}
\begin{align}
U_{2}^{c} \circ D_{2}^{c}(\widetilde{\mathbf{F}}) & = T_{\widetilde{m},\widetilde{n}}U_2D_2(T_{-\widetilde{m},-\widetilde{n}}(\widetilde{\mathbf{F}})) \\
& = T_{\widetilde{m},\widetilde{n}}U_2D_2(T_{s_x-\widetilde{m},s_y-\widetilde{n}}(\mathbf{F})) \\
& = T_{\widetilde{m},\widetilde{n}}T_{s_x-1,s_y-1}U_2D_2(T_{1-\widetilde{m},1-\widetilde{n}}(\mathbf{F})) \\
& = T_{s_x,s_y}(T_{m,n}U_2D_2(T_{-m,-n}(\mathbf{F})))  
\end{align}
\label{eq19}
\end{subequations}
where the properties $\widetilde{m} = 1 - m$ and $\widetilde{n} = 1 - n$ (for odd $s_x$ and $s_y$) are used in Eq. \ref{eq19}d and holds based on the fact that the weights of the corresponding components before and after the translation are the same. This is ensured by the global average pooling layer in CAPD as pointed out by \cite{APS,LPS}. Then, Eq. \ref{eq16} is shown to be valid by substituting Eq. \ref{eq18} into Eq. \ref{eq19}d. The same conclusion can similarly be reached when $s_x$ or $s_y$ is even, since $\widetilde{m} = m$ or $\widetilde{n}=n$ corresponds to them. 
\par 
In practice, the full shift equivalence of the network cannot be satisfied because the boundaries of the image change unpredictably before and after the input translation as shown in Fig. \ref{fig3}. Therefore, the AW module is designed to minimize the effect of boundary variations on shift equivalence. In addition, although the $T$ in Eq. \ref{eq8} closer to 0 favours shift equivalence, a higher $T$ facilitates the fusion of component features and thus improves segmentation performance. Thus, $T$ is set as a hyperparameter in this paper to balance shift equivalence and segmentation performance. 
\subsection{Loss Function}
\label{Loss Function}
A loss function that combines the cross-entropy loss $l_{ce}$ and the Dice loss $l_{de}$ was utilized. Mathematically, the loss function can be expressed as the sum of both losses, denoted as $l = l_{ce}+l_{de}$. The values of cross-entropy loss $l_{ce}$ and Dice loss $l_{de}$ for a given sample image are represented as follows:
\begin{equation}
\label{eq20}  
 l_{ce}(\hat{\mathbf{Y}},\mathbf{Y})= -\frac{1}{HW}\sum_{i=1}^{H}\sum_{j=1}^{W}\log q(x_{ij},y_{ij}) 
\end{equation}
\begin{equation}
\label{eq21}  
  l_{de}(\hat{\mathbf{Y}},\mathbf{Y}) = 1 - 2\frac{\left| \hat{\mathbf{Y}}\cap \mathbf{Y}  \right|}{\left|\hat{\mathbf{Y}}\right| + \Big|\mathbf{Y} \Big| }
\end{equation}
where $q(x_{ij},y_{ij})$ denotes the probability that the pixel $x_{ij}$ is predicted to be the ground truth $y_{ij}$. The meanings of $\hat{\mathbf{Y}}$ and $\mathbf{Y}$ are consistent with those in Eq. \ref{eq4}.
\section{EXPERIMENTS}
In this section, the dataset utilized in the experiments is first described and details on the generation of the training and test datasets are provided. The metrics employed to evaluate shift equivalence and segmentation performance are then defined. Subsequently, the shift equivalence problem for the most advanced image segmentation networks is investigated. Six networks explicitly designed to address image shift equivalence are then compared, demonstrating the efficacy of the proposed method. Following that, the effect of boundary variations is analyzed and ablation experiments are conducted. Model complexity and runtime are also further analyzed after the ablation experiments. Lastly, four other real industrial datasets are used to validate the effectiveness of the proposed method. 
\subsection{Generation of training and test datasets}
\label{Dataset}
\begin{figure}[htp]
\centering
\subfloat[]{\includegraphics[width=0.33\columnwidth]{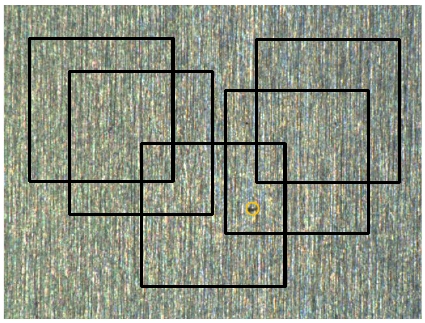}%
\label{fig7_a}}
\hfil
\subfloat[]{\includegraphics[width=0.33\columnwidth]{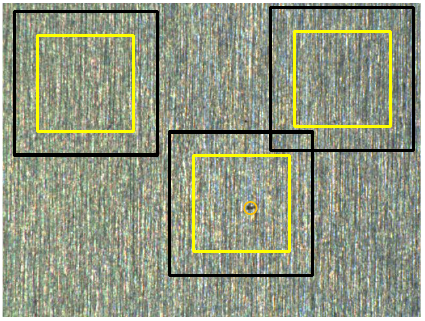}%
\label{fig7_b}}
\hfil
\subfloat[]{\includegraphics[width=0.33\columnwidth]{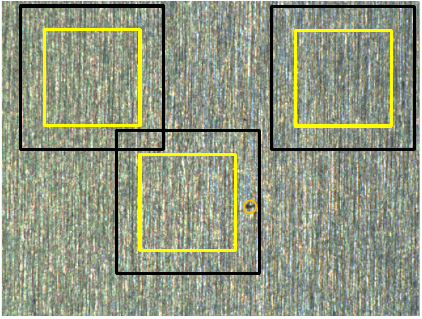}%
\label{fig7_c}}
\caption{The three sampling methods for generating dataset. (a) random sampling for the training and validation dataset. (b) sliding sampling for the MDT. (c) sliding sampling for the BDT.}
\label{fig7}
\end{figure}

A publicly available micro surface defect (MSD) of silicon steel strip dataset was used in the experiments \cite{Song}. The dataset consists of 35 images of surface defects in silicon steel strips, each with a resolution of 640×480. The defects are categorized into two groups: spot-defect images (SDI) and steel-pit-defect images (SPDI), containing 20 and 15 images, respectively. Notably, one distinctive characteristic of this dataset is the presence of random background textures in the original images, with the defects occupying a small portion of the overall image, as depicted in Fig. \ref{fig1}. The original dataset was divided in a ratio of 0.8, 0.1 and 0.1 to be used in training, validation and testing phases, respectively. Given that micro-defects are relatively sparse in comparison to the overall image, small resolution images (128×128) were cropped from the original images for the generation of training, validation and test sets. A random sampling strategy where each raw image was sampled into 30 images was employed for training and validation datasets, while maintaining a 3:1 ratio between defective and normal images, as illustrated in Fig. 7\subref{fig7_a}. Two test sets were constructed to evaluate the segmentation performance and shift equivalence, respectively:  
\par  
\textbf{Middle Defect Testset (MDT).} The MDT aims to evaluate the network when defects are located in the middle region of the image. To generate the MDT, sampling windows were moved across the images with a one-pixel increment, and only images with defects located within the yellow window were selected from the black window as shown in Fig. 7\subref{fig7_b}. The distance between the black and yellow window boundaries was set to 40 pixels. 
\par 
\textbf{Boundary Defect Testset (BDT).} The BDT was created to assess the network when defects are positioned in the boundary regions of the image. The generation of the BDT followed a similar process to the MDT, but only included images where defects appeared between the black and yellow windows, as depicted in Fig. 7\subref{fig7_c}.
\par
\begin{figure}[htp]
\centering
\subfloat[]{\includegraphics[width=0.44\columnwidth]{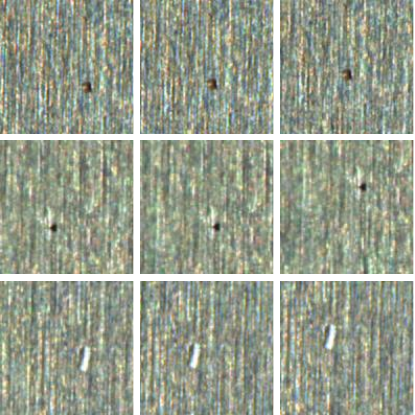}%
\label{fig8_a}}
\hfil
\subfloat[]{\includegraphics[width=0.44\columnwidth]{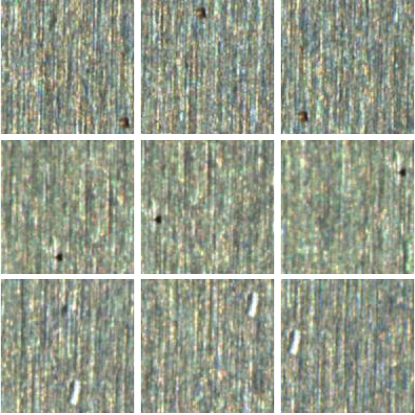}%
\label{fig8_b}}
\caption{Visualization of two test sets. (a) visualization of the MDT. (b) visualization of the BDT}
\label{fig8}
\end{figure}

 
The visualization results of the MDT and BDT can be observed in Fig. 8\subref{fig8_a} and Fig. 8\subref{fig8_b}, respectively.

\subsection{Implementation Details}
Our experiments were conducted using the PyTorch deep learning library \cite{Pytorch}. The network was optimized using the SGD optimizer \cite{SGD} with an initial learning rate of 0.001 and a momentum value of 0.9. To further improve the training performance, a polynomial learning rate scheduling approach with a power of 0.9 was employed in the experiments. The training process was carried out for a maximum of 500 epochs, with a batch size of 32, utilizing an NVIDIA GeForce RTX 3090 GPU. The training phase continued until there was no decrease in the loss of the validation set for 10 consecutive epochs.
\subsection{Evaluation Metrics}
To assess the shift equivalence in our experiments, we designed two new metrics, namely mean variance of Intersection-over-Union (mvIoU) and mean variance of defect area (mvda). Both metrics are designed to describe fluctuations in defect segmentation masks. A lower value for mvIoU and mvda indicates a higher level of shift equivalence. Additionally, we utilized the mean Intersection-over-Union (mIoU), precision, recall and f1-score as measures of segmentation performance. Let us consider the test set, whether it is the MDT or BDT, divided into $N$ subsets: $M_{j},\ j=1,2,\dots,N$, each subset consists of images cropped from the same raw image. $IoU(\hat{\mathbf{Y}}_i,\mathbf{Y}_i)$ denotes the IoU between the predicted segmentation $\hat{\mathbf{Y}}_i$ and the ground truth $\mathbf{Y}_i$ corresponding to input image $\mathbf{X}_i$, as described in Eq. \ref{eq4}. The equivalence metrics mvIoU and mvda are defined as follows:
\par 
$\mathbf{mvIoU}$: The mIoU of the set $M_j$ is calculated as:
\begin{equation}
\label{eq22}  
 mIoU_{j} = \frac{1}{\left|M_j\right|}\sum_{i=1,\mathbf{X}_i \in M_j}^{\left|M_j\right|}IoU(\hat{\mathbf{Y}}_i,\mathbf{Y}_i)
\end{equation}
\par 
The metric mvIoU which portrays the equivalence of segmentation masks is formulated as:
\begin{equation}
\label{eq23}  
 \mathrm{mvIoU} = \frac{1}{N}\sum_{j=1}^{N}\frac{1}{\left|M_j\right|-1}\sum_{i=1,\mathbf{X}_i \in M_j}^{\left|M_j\right|}(IoU(\hat{\mathbf{Y}}_i,\mathbf{Y}_i)-mIoU_j)^2
\end{equation}
\par 
$\mathbf{mvda}$: Assume that the area of defects in $\mathbf{X}_i$ is $Area(\mathbf{X}_i)$, then the average area of the predicted defects in the set $M_j$ can be expressed as:
\begin{equation}
\label{eq24}  
 mArea_{j} = \frac{1}{\left|M_j\right|}\sum_{i=1,\mathbf{X}_i \in M_j}^{\left|M_j\right|}Area(\mathbf{X}_i)
\end{equation}
\par 
The metric mvda is calculated as:
\begin{equation}
\label{eq25}  
 \mathrm{mvda} = \frac{1}{N}\sum_{j=1}^{N}\frac{1}{\left|M_j\right|-1}\sum_{i=1,\mathbf{X}_i \in M_j}^{\left|M_j\right|}(Area(\mathbf{X}_i)-mArea_j)^2
\end{equation}
\par 
To calculate the area of defects, only the connected defect domain with the largest area in the segmentation masks is considered, and the rest is deemed as overkill in defect segmentation.
\begin{table*}[!t]
	\centering
	\begin{threeparttable}
  \caption{Comparison of Current Advanced Segmentation Networks on Shift Equivalence and Segmentation performance. }
  \centering
  \label{tab2}
  \renewcommand{\arraystretch}{1.3}
	\begin{tabular}{>{\centering\arraybackslash}m{2.2cm}*{5}{>{\centering\arraybackslash}m{1.8cm}}{>{\centering\arraybackslash}m{2.5cm}}}
	\toprule
	method      & mIoU $\uparrow$  (\%)          & precision $\uparrow$ (\%)     & recall $\uparrow$ (\%)            & f1-score $\uparrow$        & mvIoU $\downarrow$            & mvda $\downarrow$              \\  \hline
	PSPNet \cite{Pspnet}     & 69.16 / 60.74 & 79.93 / 78.47 & 83.13 / 82.70     & 0.8150 / 0.8053  & 0.0044 / 0.0178  & 55.5577 / 152.3609 \\
	UperNet \cite{Upernet}     & 73.38 / 72.13 & 81.01 / 80.79 & 85.64 / 85.83     & 0.8326 / 0.8323 & 0.0018 /  0.0038 & 45.5926 /  72.1611 \\
	DeepLabv3+ \cite{deeplabv3+}  & 73.56 / 70.87 & 83.22 / 82.01 & 86.82 / 85.18     & 0.8498 / 0.8356 & 0.0015 / 0.0039  & 34.5886 / 71.2481  \\
	Mask2former \cite{Mask2former}  & \underline{79.21} / \textbf{78.34} & \underline{86.45} / \underline{87.01} & \underline{91.18}  /   \textbf{89.34} & \underline{0.8875} / \textbf{0.8816} & 0.0012 / 0.0022  & 51.4333 / 51.0220  \\
	SAM-Adapter \cite{SAM-Adapter}  & \textbf{79.40} / \underline{78.09} & \textbf{90.35} / \textbf{90.61} & 87.63 / 85.81     & \textbf{0.8897} / \underline{0.8814} & \underline{0.0011} / \textbf{0.0020}  & \underline{18.9232} / \underline{34.1047}  \\
	Ours (CAPS) & 78.15 / 75.93 & 84.19 / 83.32 & \textbf{93.02} / \underline{87.35}     & 0.8839 / 0.8529 & \textbf{0.0010} / \underline{0.0021}  & \textbf{2.4139} / \textbf{3.5229}    \\
	\bottomrule
\end{tabular}
	       \begin{tablenotes}
	\item * The best results are shown in \textbf{bold} and the second best results are \underline{underlined}. ($x$ / $y$) refers to the values of the metrics on the MDT and BDT as $x$ and $y$, respectively.
\end{tablenotes}
\end{threeparttable}
\end{table*}
\subsection{Comparison with current advanced segmentation networks}
The current advanced segmentation network designs have not explicitly focused on shift equivalence due to their emphasis on segmentation performance. To investigate the shift equivalence of current state-of-the-art segmentation networks, five high-performing networks were implemented for evaluation:1) UperNet \cite{Upernet}: A multi-task learning framework that performs well on image segmentation by parsing multiple visual concepts such as category, material and texture; 2) PSPNet \cite{Pspnet}: A network with a pyramid pooling module designed to achieve excellent image segmentation performance by fusing features from different receptive fields; 3) DeepLabv3+ \cite{deeplabv3+}: An encoder-decoder network that utilizes the Atrous Spatial Pyramid Pooling module to extract multi-scale contextual features using dilated convolutions; 4) Mask2former \cite{Mask2former}: A network that employs a transformer decoder with masked attention, aiming to extract local features within the region of the predicted mask. It currently achieves state-of-the-art semantic segmentation performance on various publicly available datasets; 5) SAM-Adapter \cite{SAM-Adapter}: A network adapter that builds upon the Segment Anything Model \cite{SAM} as a foundation model. It incorporates multiple visual prompts to adapt to downstream tasks. In our evaluation, ResNet-101 was used as the backbone for UperNet, PSPNet, and DeepLabv3+, while employing Swin-Transformer-large as the backbone for Mask2former. All four models utilize the official code from the mmsegmentation library \footnote{mmsegmentation: \url{https://github.com/open-mmlab/mmsegmentation}} and employ the default pretrained model to achieve optimal segmentation performance. SAM-Adapter was implemented using the official code \footnote{SAM-Adapter: \url{https://github.com/tianrun-chen/SAM-Adapter-PyTorch}} initialized with SAM-Large pretrained parameters for feature extraction.
\par 
Table \ref{tab2} presents the results of five advanced image segmentation networks and ours, on the MDT and BDT. It is worth noting that our network, depicted in Fig. \ref{fig4} is relatively more lightweight compared to the others, which introduces some unfairness in the comparison. However, our network still exhibits superior equivalence, particularly in terms of the mvda. The segmentation performance surpasses that of DeepLabv3+, indicating that the proposed method can simultaneously balance shift equivalence and segmentation performance. It is observed that most existing segmentation networks suffer from low shift equivalence, so it is crucial to explore methods for improving equivalence in both academic research and real-world industrial applications. 
\subsection{Comparison with other advanced shift equivalence methods}
The proposed CAPS method was compared with six advanced methods aiming to enhance shift equivalence, namely BlurPool \cite{BlurPool}, APS \cite{APS},  LPS \cite{LPS}, PBP \cite{PBP}, MWCNN \cite{MWCNN}, and DUNet \cite{DUNet} on both the MDT and BDT. To ensure experimental fairness, all methods except DUNet utilized the U-Net structure depicted in Fig. \ref{fig4} as a base model, with only the downsampling and upsampling layers replaced. Unlike other methods, DUNet \cite{DUNet} replaces only a portion of the standard convolutions in U-Net with deformable convolutional blocks. To adhere to the recommendation of APS and LPS, circular padding was utilized in all experiments, while keeping all other settings consistent with their original papers and codes \footnote{BlurPool: \url{https://github.com/adobe/antialiased-cnns}}  \footnote{APS: \url{https://github.com/achaman2/truly_shift_invariant_cnns}} \footnote{LPS: \url{https://raymond-yeh.com/learnable_polyphase_sampling}} \footnote{PBP: \url{https://github.com/Moshtaf/shift-invariant-unet}} \footnote{MWCNN: \url{https://github.com/lpj0/MWCNN}} \footnote{DUNet: \url{https://github.com/RanSuLab/DUNet-retinal-vessel-detection}}.  
\par 
Table \ref{tab3} provides a comparison of different methods that contribute to the improvement of shift equivalence on the MDT and BDT. The best results are shown in bold and the second best results are underlined for a clearer comparison. CAPS greatly reduces the mvIoU and mvda on both test sets, revealing better shift equivalence. Concretely, CAPS relatively reduces upon the second best method by 23.08\% mvIoU, 70.28\% mvda on the MDT and 12.50\% mvIoU, 82.32\% mvda on the BDT. The improvement in equivalence reveals the importance of considering variations in feature boundaries during the downsampling process. The results on the BDT substantially surpass those of other methods, further indicating that the AW module does not negatively impact segmentation performance and equivalence when defects are located at the image boundaries. Apart from the improvement in shift equivalence, CAPS achieve a new state-of-the-art of 75.15\%, 75.93\% mIoU, surpassing the previous best solution PBP by +0.55\% and +0.36\% on the MDT and BDT, respectively. Although CAPS does not reach optimality in term of precision, it exhibits +5.04\% and +1.34\% recall improvement compared with the second best method LPS. The optimal values obtained on the f1-score (0.8839 on the MDT and 0.8529 on the BDT) also demonstrate that CAPS has superior segmentation performance while balancing precision and recall. It can also be observed that DUNet and MWCNN exhibit poor shift equivalence and segmentation performance, even worse than Baseline. 
\par 
\begin{figure}[htp]
\centering
\includegraphics[width=\columnwidth]{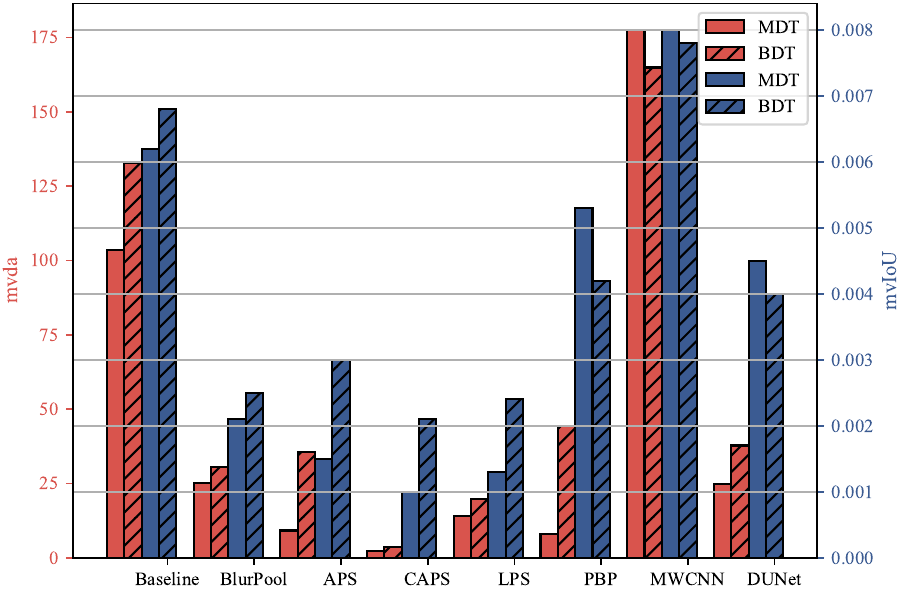}
\caption{Comparison of mvda and mvIoU between the MDT and BDT.}
\label{fig9}
\end{figure}

\begin{table*}[!t]
		\centering
	\begin{threeparttable}
	\caption{Comparison with other advanced shift equivalence methods on shift equivalence and segmentation performance.}
	\centering
	\label{tab3}
	\renewcommand{\arraystretch}{1.3}
	\begin{tabular}{>{\centering\arraybackslash}m{2cm}*{5}{>{\centering\arraybackslash}m{1.8cm}}{>{\centering\arraybackslash}m{2.5cm}}}
		\toprule
	method   & mIoU $\uparrow$ (\%)          & precision $\uparrow$ (\%)     & recall $\uparrow$ (\%)        & f1-score $\uparrow$        & mvIoU $\downarrow$           & mvda $\downarrow$               \\   \hline
	Baseline \cite{unet} & 69.84 / 68.65 & 83.37 / 81.49 & 83.07 / 82.24 & 0.8322 / 0.8186 & 0.0062 / 0.0068 & 103.4031 / 132.9349 \\
	BlurPool \cite{BlurPool} & 73.16 / 73.35 & 90.06 / 81.61 & 80.15 / 80.63 & 0.8482 / 0.8112 & 0.0021 / 0.0025 & 25.3123 / 30.6426   \\
	APS \cite{APS}     & 68.41 / 69.26 & 89.22 / 80.37 & 76.82 / 77.71 & 0.8256 / 0.7902 & 0.0015 / 0.0030  & 9.1908 / 35.6342    \\
	LPS \cite{LPS}     & 72.04 / 72.83 & \underline{84.47} / 82.24 & \underline{87.98} / \underline{86.01} & 0.8619 / \underline{0.8408} & \underline{0.0013} / \underline{0.0024} & 14.0403 / \underline{19.9224}   \\
	PBP \cite{PBP}     & \underline{77.60} / \underline{75.57} & \textbf{94.01} / 82.47 & 82.18 / 81.15 & \underline{0.8770} / 0.8180  & 0.0053 / 0.0042 & \underline{8.1232} / 44.2956    \\
	MWCNN \cite{MWCNN}    & 63.64 / 66.35 & 76.24 / 80.88 & 84.61 / 82.44 & 0.8021 / 0.8165 & 0.0080 / 0.0078 & 177.5254 / 164.8511 \\
	DUNet \cite{DUNet}   & 62.42 / 67.24 & 81.39 / \textbf{84.17} & 75.57 / 76.16 & 0.7837 / 0.7996 & 0.0045 / 0.0040  & 24.8303 / 37.7710   \\
	CAPS     & \textbf{78.15} / \textbf{75.93} & 84.19 / \underline{83.32} & \textbf{93.02} / \textbf{87.35} & \textbf{0.8839} / \textbf{0.8529} & \textbf{0.0010} / \textbf{0.0021}  & \textbf{2.4139} / \textbf{3.5229}    \\
	\bottomrule 
	\end{tabular}
		       \begin{tablenotes}
		\item * The best results are shown in \textbf{bold} and the second best results are \underline{underlined}. ($x$ / $y$) refers to the values of the metrics on the MDT and BDT as $x$ and $y$, respectively. 
	\end{tablenotes}
	\end{threeparttable}
\end{table*}

\textbf{Comparison of equivalence between the MDT and BDT.} Fig. \ref{fig9} shows the comparison of mvda and mvIoU between the MDT and BDT. The red bar represents the value of mvda and corresponds to the y-axis on the left, while the blue bar represents mvIoU and corresponds to the y-axis on the right. It is evident that the baseline method exhibits lower equivalence compared to the other methods, emphasizing the importance of redesigning the downsampling and upsampling structure. The results illustrate that almost all methods achieve higher values of mvda and mvIoU on the BDT dataset compared to the MDT dataset, indicating the increased difficulty in maintaining equivalence when defects are located at the boundaries rather than the middle region.
\par
\begin{figure*}[!bt]
\centering
\includegraphics[width=1.8\columnwidth]{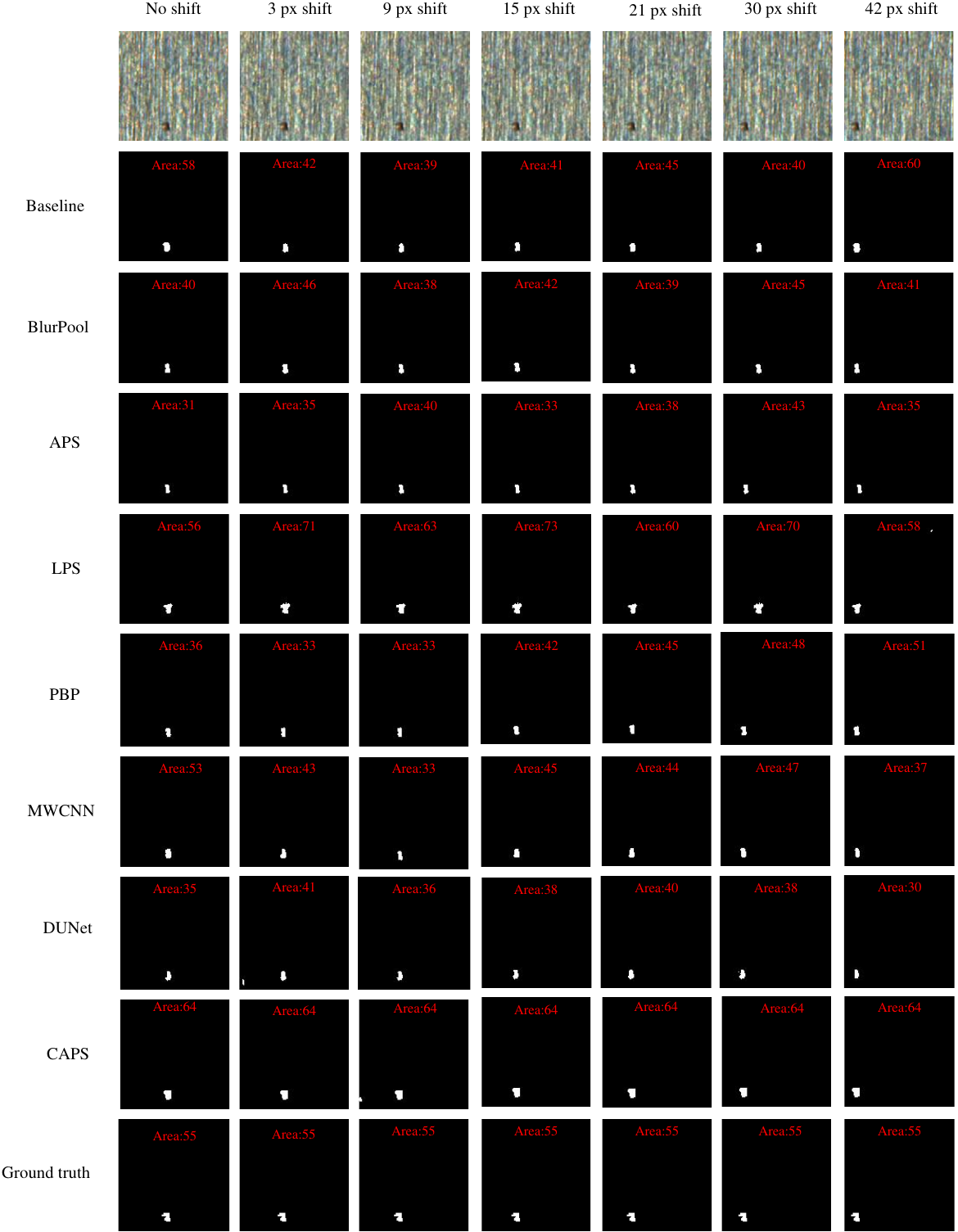}
\caption{The qualitative segmentation results of the different methods on the BDT. The area of the defect is labeled at the top of the image using red font. Compared to other methods, the segmentation masks of CAPS have the least fluctuation and possess the best shift equivalence.}
\label{fig10}
\end{figure*}

 
\textbf{Qualitative result analysis of different methods.} The qualitative segmentation results of the different methods are shown in Fig. \ref{fig10}, using the more challenging BDT. The first row describes the result of original image without any shift, while each successive row is shifted to the left by a specified number of pixels. Compared to other methods, CAPS demonstrates nearly complete equivalence in the segmentation results, with the exception of a slight discrepancy observed when shifting by 9 pixels, as depicted in the ninth row. Conversely, other methods, such as BlurPool, exhibit significant fluctuations in the segmentation mask, particularly in terms of the area of defects.
\begin{figure}[H]
\centering
\includegraphics[width=0.95\columnwidth]{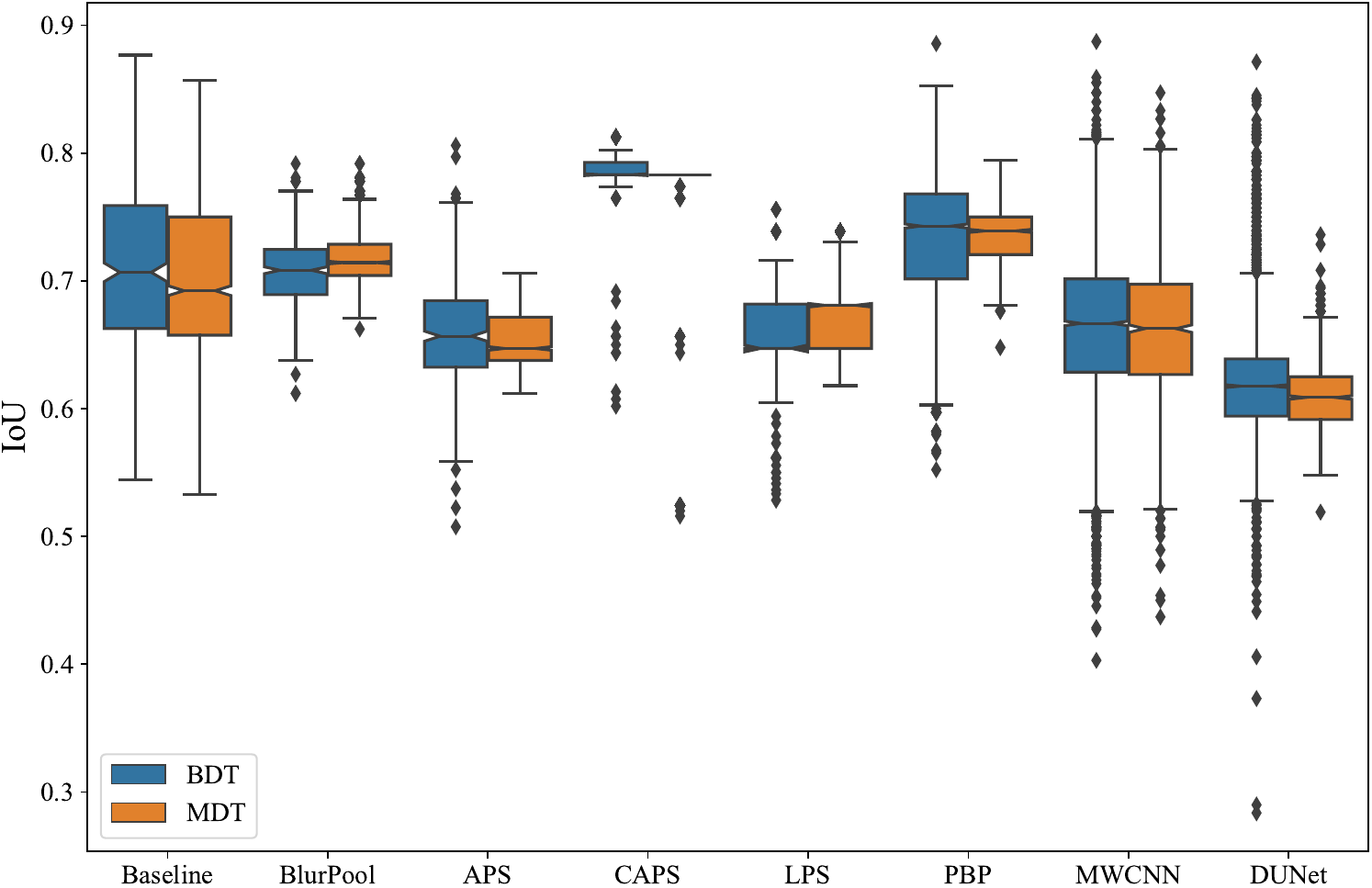}
\caption{The IoU results when the raw image $\textit{SDI\_3}$ was sampled.}
\label{fig11}
\end{figure}
\subsection{Effect of boundary variations on shift equivalence}
To clarify the reasons for the poor performance of shift equivalence on the BDT, the images sampled from a raw image named $\textit{SDI\_3}$ were taken out individually to test their IoU. Fig. \ref{fig11} shows the results in the form of box plots, where the height of the boxes indicates the extent of IoU fluctuation, and the black dots signify outliers. Notably, the IoU exhibits more outliers when the defects lie at the image boundaries. This suggests that the main reason for the weaker equivalence on the BDT than on the MDT is that the segmentation results on image boundaries are more susceptible to boundary variations due to translations. Therefore, some outliers in Fig. \ref{fig11} are more likely to be generated on the BDT, affecting the equivalence of the network. 
\par 
\begin{figure}[]
	\centering
	\includegraphics[width=\columnwidth]{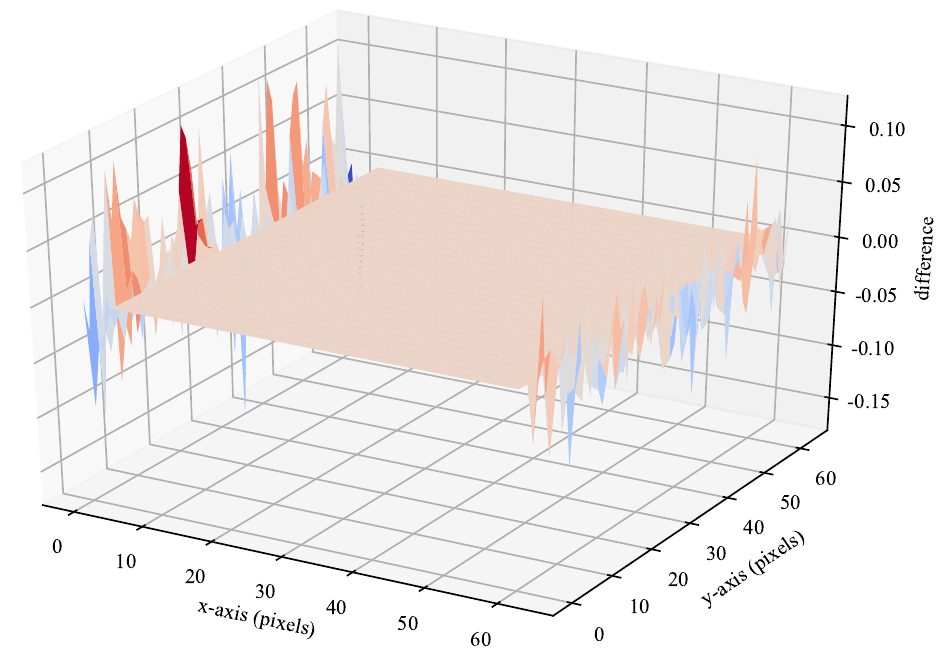}  %pic11.pdf
	\caption{Boundary differences due to translation of the same component feature map.}
	\label{fig12}
\end{figure}
Fig. \ref{fig11} illustrates the boundary differences in downsampled features that arise from translation. Specifically, the input to the proposed network consists of two sampled images, one with a black box and the other with a red box shown in Fig. \ref{fig1}. The latter is obtained by translating the former down one pixel. Assuming that the first channels in the feature maps of these two images after the first CAPD layer are denoted as $\mathbf{D}_1$ and $\mathbf{D}_2$  with a resolution of $64\times64$. The difference in the z-axis of Fig. \ref{fig12} can be calculated as: $Shift(\mathbf{D}_1) - \mathbf{D}_2$. The middle pink area represents that the two features are identical before and after translation, but the boundary region features exhibit significant differences. These boundary differences introduce uncertainty in the component fusion process of CAPS. So the AW module is designed to disregard feature boundaries shown in Fig. \ref{fig12} when generating the weights for component fusion. By doing so, the downsampling results are similar when the input image is shifted, which improves the shift equivalence.

\subsection{Ablation analysis}

\begin{table*}[t]
	\centering
	\begin{threeparttable}
	\caption{THE RESULTS OF ABLATION EXPERIMENTS}
	\label{tab4}
	\renewcommand{\arraystretch}{1.7}
	\begin{tabular}{*{4}{>{\centering\arraybackslash}m{0.5cm}}*{5}{>{\centering\arraybackslash}m{1.8cm}}*{1}{>{\centering\arraybackslash}m{2.1cm}}}
		\toprule
	AW & CA & LPF & DA & mIoU $\uparrow$ (\%)           & precision $\uparrow$ (\%)     & recall $\uparrow$ (\%)         & f1-score $\uparrow$       & mvIoU $\downarrow$          & mvda $\downarrow$             \\ \hline
	& $\surd$  & $\surd$   &    & 71.63 / 69.50  & 83.90 / 81.98 & 82.60 / 82.30 & 0.8324 / 0.8214 & 0.0015 / 0.0027 & 15.2719 / 20.0064 \\
	$\surd$  &    &$\surd$   &    & 75.17 /  72.50 & 82.21 / 80.24 & 87.26 / 86.01 & 0.8466 / 0.8302 & 0.0012 / 0.0022 & 7.6788 / 13.3844  \\
	$\surd$ &$\surd$  &     &    & 72.28 / 72.21  & 77.03 / 78.65 & \underline{92.75} / 87.09 & 0.8416 / 0.8266 & \textbf{0.0006} / \textbf{0.0019} & \textbf{0.0002} / \textbf{0.3806}     \\
	$\surd$  &$\surd$  &$\surd$   &    & \underline{78.15} / \underline{75.93}  & \underline{84.19} / \underline{83.32} & \textbf{93.02} / \underline{87.35} & \textbf{0.8839} / \underline{0.8529} &\underline{0.0010} / \underline{0.0021}  & \underline{2.4139} / \underline{3.5229}   \\
	$\surd$  &$\surd$  & $\surd$   &$\surd$  & \textbf{78.23} / \textbf{76.01}  & \textbf{84.35} / \textbf{83.41} & 92.11 / \textbf{87.43} & \underline{0.8806} / \textbf{0.8537} & 0.0021 / 0.0033 & 3.5132 / 3.9946  \\
		\bottomrule
	\end{tabular}
	       \begin{tablenotes}
		\item * The best results are shown in \textbf{bold} and the second best results are \underline{underlined}. ($x$ / $y$) refers to the values of the metrics on the MDT and BDT as $x$ and $y$, respectively. 
	\end{tablenotes}
	\end{threeparttable}
\end{table*}

\begin{figure*}[b]
	\centering
	\subfloat[]{\includegraphics[width=0.9\columnwidth]{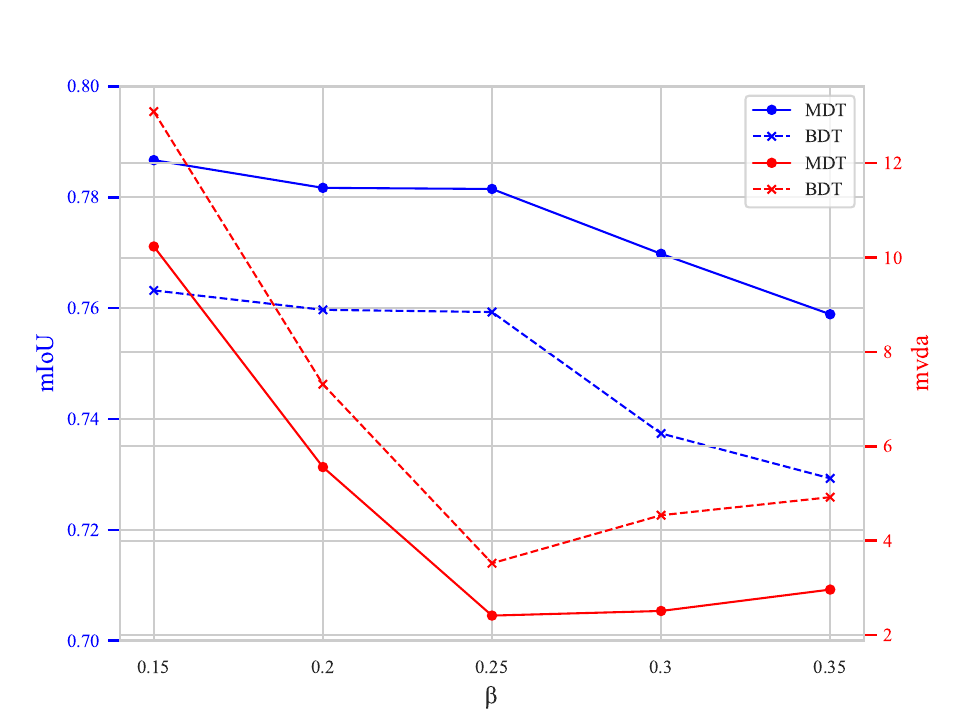}%
		\label{fig13_a}}
	\hfil
	\subfloat[]{\includegraphics[width=0.9\columnwidth]{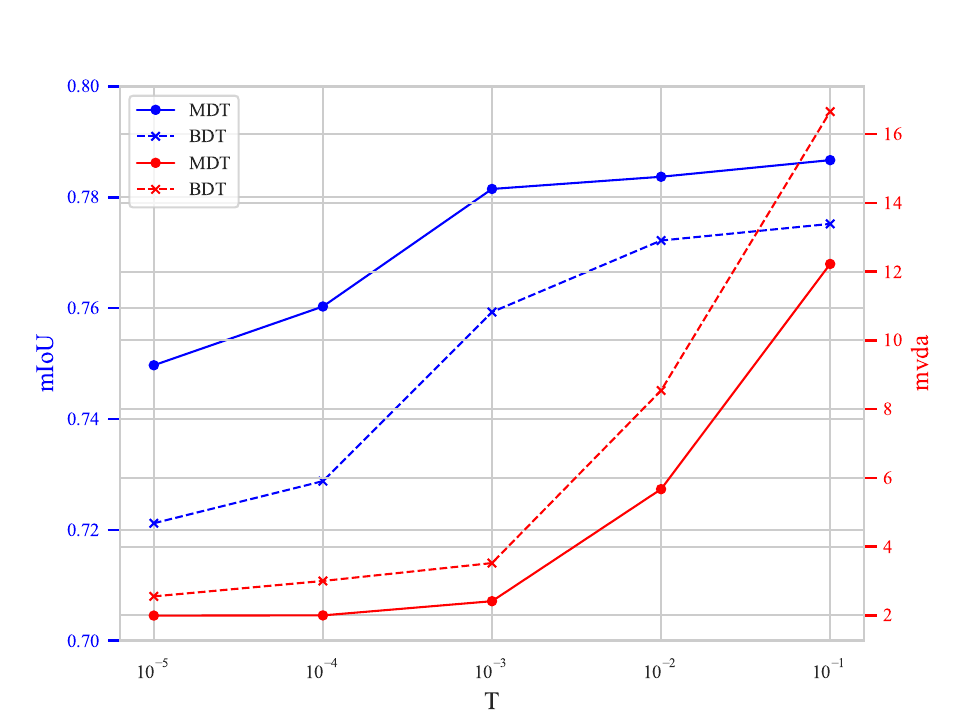}%
		\label{fig13_b}}
	\caption{Sensitivity analysis of the hyperparameter. (a) the analysis of $\beta$. (b) the analysis of $T$.}
	\label{fig13}
\end{figure*}
In this subsection, we conduct an ablation analysis on the collected MDT and BDT to verify the effectiveness of modules designed in CAPS and analyze the hyperparameters in the proposed methods. Specifically, the efficacy of the AW, CA and LPF in CAPS is assessed. In addition to this, the effectiveness of data augmentation (DA) is further evaluated on CAPS by applying random transformations to the training data, including random rotation, flip, brightness, contrast and cutout \cite{cutout}.
\par 
\textbf{Effect of the AW:} Comparing the first and fourth rows in Table \ref{tab4}, it can be seen that the removal of AW module leads to a relative improvement of $50.0\%$ and $28.6\%$ in terms of mIoU on the MDT and BDT, respectively, while mvda increases by 4 to 6 times. This demonstrates the crucial role of the AW module in achieving shift equivalence. Furthermore, a substantial decrease in mIoU is observed when the AW module is removed. This overall reduction in segmentation performance highlights the significance of the AW module in maintaining segmentation performance.
\par 
\textbf{Effect of the CA:} As depicted in the second row of Table \ref{tab4}, the removal of the CA module leads to a relative decrease on mIoU by $3.8\%$ and $4.5\%$ on the MDT and BDT, respectively. Furthermore, the shift equivalence of the network decreases, as indicated by the increase on mvIoU and mvda. This emphasizes the effectiveness of utilizing means based on the fusion of attentional components, not only in enhancing the segmentation performance of the network but also in improving shift equivalence. 
\par 
\textbf{Effect of the LPF:} As shown in the third and fourth rows of Table \ref{tab4}, the use of LPF effectively improves segmentation performance with +5.87\% mIoU, +7.16\% precision on the MDT and +3.72\% mIoU, +4.67\% precision on the BDT. However, shift equivalence is compromised, as indicated by the increase in both mvIoU and mvda on the MDT and BDT. This decrease in equivalence can be attributed to the LPF further blurring boundary features, leading to increased variations at the feature boundaries before and after translation. Therefore, it is suggested that when a higher demand for equivalence is prioritized over segmentation performance, the removal of LPF in CAPS can contribute to an increase in shift equivalence.
\par
\textbf{Effect of the DA:} It can be seen from Table \ref{tab4}, there is a slight increase in segmentation performance (e.g. from 78.15\% to 78.23\% in mIoU) but a decrease in shift equivalence (e.g. from 2.4139 to 3.5132 in mvda) on the MDT. It can be seen that data augmentation enhances the diversity of samples, thus benefiting the segmentation performance, but it cannot accurately improve the network’s shift equivalence. Sometimes the distribution between the augmented training data and the original test data is biased, which reduces the equivalence of the network.
\par
\textbf{Effect of the hyperparameters:} Two main hyperparameters are used in our method: the $\beta$, which controls the proportion of windowing, and the $T$ when the components are fused. We investigated the impact of varying $\beta$ and $T$ within a certain range on mIoU and mvda as depicted in Fig. \ref{fig13}. The blue y-axis on the left represents mIoU, while the red y-axis on the right represents mvda. The solid and dashed lines in both images depict the experimental results on the MDT and BDT, respectively.
\par

Specifically, in Fig. 13\subref{fig13_a} , the $\beta$  indicates the truncation ratio of the AW module to the feature boundaries, which means that higher values make downsampling less affected by image boundaries. It can be found that the network exhibits the best shift equivalence for a $\beta$ value of 0.25, with the mIoU only slightly lower than the highest value. Therefore, the hyperparameter $\beta$ was consistently set to 0.25 in all experiments. As shown in Fig. 13\subref{fig13_b}, the temperature control factor $T$ determines how well the component features are fused. When $T$ approaches 0, the \textit{T-softmax} function approximates the \textit{argmax} operation, thereby increasing the shift equivalence for smaller mvda no matter on the MDT and BDT. Conversely, as $T$ approaches 1, the \textit{T-softmax} function becomes equivalent to the standard \text{softmax} function, enhancing component fusion, which benefits improving segmentation performance. However, when $T$ exceeds $10^{-3}$ the equivalence of the segmentation network drops drastically, as shown by the red solid and dashed lines in Fig. 13\subref{fig13_b}. To strike a balance between segmentation performance and shift equivalence we set $T$ to $10^{-3}$ to achieve this trade-off. 

\subsection{Model complexity and runtime analysis}
In order to check the complexity of our CAPS, the number of parameters and FLOPs are analysed in the proposed method and compared with other advanced methods as shown in Table \ref{tab5}. The average inference time for a single image is illustrated in Table \ref{tab6}. Although the number of model parameters as well as FLOPs is larger than several methods, the inference time for a single image is within the requirements of real industrial scenarios ($\le$40 ms). Specifically, the average inference time for a single image is 9.24 ms, 13.33 ms and 38.93 ms when the input size is 128×128, 256×256 and 512×512, respectively. Moreover, the proposed CAPS has the best shift equivalence among all the methods, so the appropriate increment in inference time compared to other methods is acceptable.
\begin{table}[H]
	\caption{Analysis of model complexity}
	\centering
	\label{tab5}
	\renewcommand{\arraystretch}{1.1}
	\begin{tabular}{>{\centering\arraybackslash}m{2cm}*{2}{>{\centering\arraybackslash}m{1.2cm}}}
		\toprule
		method   & \#params (M) & FLOPs (G) \\   \hline
		Baseline \cite{unet} & 31.04       & 13.69                          \\
		BlurPool \cite{BlurPool} & 31.04       & 13.69                        \\
		APS \cite{APS}     & 31.04       & 13.69                       \\
		LPS \cite{LPS}     & 40.79       & 21.20                          \\
		PBP \cite{PBP}     & 31.04       & 13.69                        \\
		MWCNN \cite{MWCNN}   & 45.49       & 13.95                          \\
		DUNet \cite{DUNet}   & 31.23       & 14.64                        \\
		CAPS     & 40.79       & 21.20                          \\
		\bottomrule              
	\end{tabular}
\end{table}

\begin{table}[H]
		\caption{Average inference time for a single image at different scales}
	\centering
	\label{tab6}
	\renewcommand{\arraystretch}{1.1}
	\begin{tabular}{>{\centering\arraybackslash}m{2cm}*{3}{>{\centering\arraybackslash}m{1.2cm}}}
		\toprule
	method	& 128×128 (ms) & 256×256 (ms) & 512×512 (ms) \\    \hline
		Baseline \cite{unet}& 3.71    & 5.89    & 22.67   \\
		BlurPool \cite{BlurPool}& 4.27    & 6.57    & 25.00   \\
		APS \cite{APS}     & 13.10   & 30.65   & 103.81  \\
		LPS  \cite{LPS}    & 8.08    & 12.60   & 38.78   \\
		PBP  \cite{PBP}    & 4.52    & 6.31    & 26.85   \\
		MWCNN \cite{MWCNN}   & 7.10    & 10.71   & 35.22   \\
		DUNet \cite{DUNet}   & 26.36   & 75.43   & 297.81  \\
		CAPS     & 9.24    & 13.33   & 38.93   \\
			\bottomrule      
	\end{tabular}
\end{table}

\subsection{Performance in other real-world industrial defect datasets}
\begin{figure}[H]
	\centering
	\subfloat[]{\includegraphics[width=0.4\columnwidth]{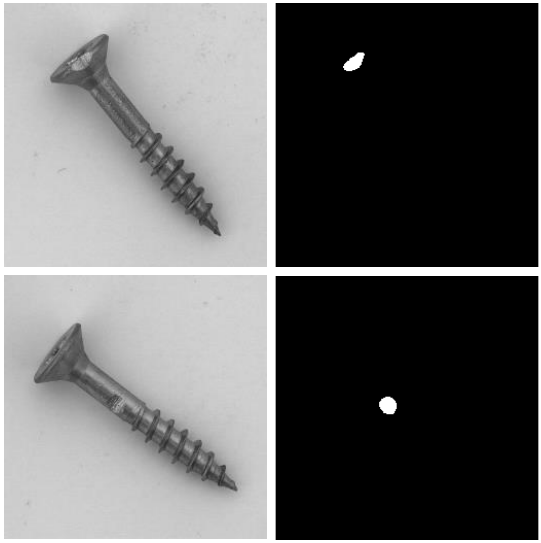}%
		\label{fig14_a}}
	\hfil
	\subfloat[]{\includegraphics[width=0.4\columnwidth]{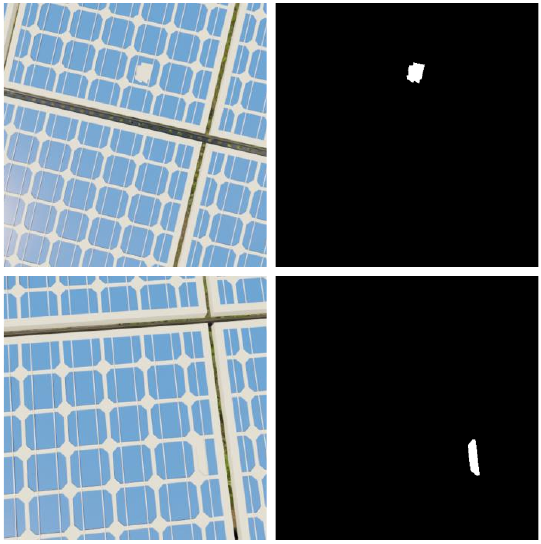}%
		\label{fig14_b}}
	
	\subfloat[]{\includegraphics[width=0.4\columnwidth]{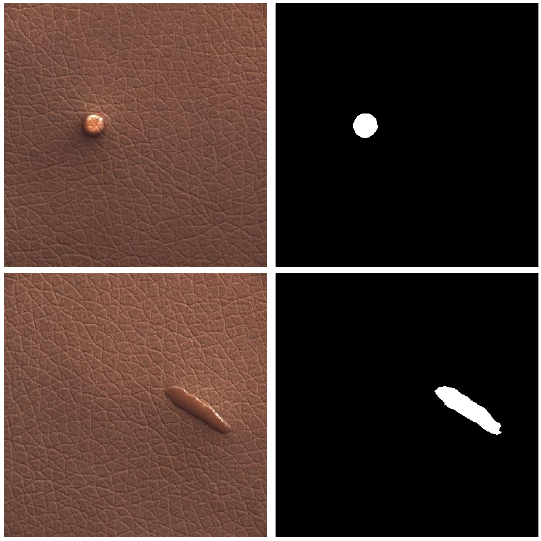}%
		\label{fig14_c}}
	\hfil
	\subfloat[]{\includegraphics[width=0.4\columnwidth]{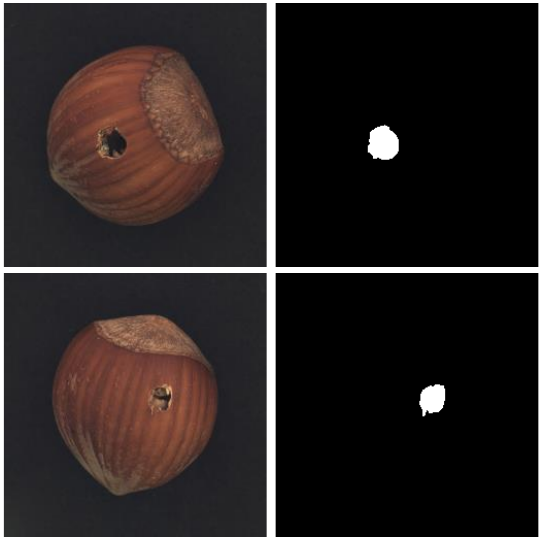}%
		\label{fig14_d}}
	\caption{Visualization of four real industrial datasets. (a) screw  (b) photovoltaic modules (c) leather (d) hazelnut}
	\label{fig14}
\end{figure}

To validate the effectiveness of the proposed method, four additional datasets were used to further evaluate the performance. Specifically, they are screw, leather and hazelnut from MVTec Anomaly Detection Dataset (MVTec AD) \cite{mvtec} and photovoltaic modules from Maintenance Inspection Dataset (MIAD) \cite{MIAD}. Fig. \ref{fig14} shows the original image and ground truth for sample images from different datasets. The number of images in each dataset and the size of the original images are shown in second and third columns of Table \ref{tab7}. The training, validation and test sets are generated in line with MSD as shown in \ref{Dataset}. Additionally, defects located both in the middle and the boundaries are collectively tested to assess the overall performance of different methods. 
\par  
\begin{table}[H]
	\caption{Details of the datasets}
	\centering
	\label{tab7}
	\renewcommand{\arraystretch}{1.4}
	\begin{tabular}{>{\centering\arraybackslash}m{1.5cm}*{4}{>{\centering\arraybackslash}m{1.2cm}}}
		\toprule
		dataset              & number of pictures & original shape & resize shape & crop size \\   \hline
		screw                & 119                & 1024×1024     & 256×256      & 128×128   \\
		leather              & 92                 & 1024×1024     & 384×384      & 128×128   \\
		hazelnut             & 70                 & 1024×1024     & 256×256      & 128×128   \\
		photovoltaic modules & 2500               & 512×512       & 384×384      & 128×128   \\
		\bottomrule
	\end{tabular}
\end{table}

\begin{table*}[!bt]
		\centering
	\begin{threeparttable}
	\caption{Comparison of performance in real-world industrial defect datasets.}
	\centering
	\label{tab8}
	\renewcommand{\arraystretch}{1.1}	
	\begin{tabular}{>{\centering\arraybackslash}m{2cm}>{\centering\arraybackslash}m{1.8cm}*{5}{>{\centering\arraybackslash}m{1.3cm}}{>{\centering\arraybackslash}m{1.5cm}}}
		\toprule
		dataset                       & method   & mIoU $\uparrow$ (\%)  & precision $\uparrow$ (\%)& recall $\uparrow$ (\%)& f1-score $\uparrow$& mvIoU $\downarrow$      & mvda $\downarrow$     \\   \hline
		\multirow{8}{*}{screw}        & Baseline \cite{unet} & 74.39 & 87.82    & 84.21 & 0.8598   & 0.0008      & 69.8171   \\
		& BlurPool \cite{BlurPool}& \underline{80.98} & 90.18    & 88.99 & 0.8958   & 0.0006      & 18.8074   \\
		& APS \cite{APS}    & 75.21 & \textbf{92.45}    & 79.65 & 0.8557   & \underline{0.0002}      & 19.4182   \\
		& LPS \cite{LPS}     & 71.32 & 80.99    & 84.88 & 0.8289   & 0.0011      & \underline{16.4240}    \\
		& PBP \cite{PBP}    & 78.32 & 89.61    & 86.69 & 0.8813   & 0.0009      & 29.7078   \\
		& MWCNN \cite{MWCNN}    & 76.61 & 84.33    & \underline{89.25} & 0.8672   & 0.0014      & 102.4718  \\
		& DUNet \cite{DUNet}    & 81.25 & 89.88    & \textbf{89.67} & \underline{0.8977}   & 0.0006      & 65.4580    \\ 
		& CAPS     & \textbf{81.84} & \underline{91.19}    & 88.70  &\textbf{ 0.8993}   & \textbf{0.0001}      & \textbf{16.2942}   \\    \hline
		\multirow{8}{*}{leather}      & Baseline \cite{unet}& 76.03 & 85.88    & 88.35 & 0.8710   & 0.0003      & 663.4814  \\
		& BlurPool \cite{BlurPool}& 78.83 & 87.83    & 87.96 & 0.8789   & 0.0002      & 358.9552  \\
		& APS \cite{APS}     & 77.86 & \textbf{93.04}    & 82.69 & 0.8756   & 0.0001      & 311.9144  \\
		& LPS \cite{LPS}     & 77.06 & 85.40     & \underline{88.45} & 0.8690   & 0.0004      & \underline{67.5913}   \\
		& PBP \cite{PBP}     & \underline{80.57} & 90.82    & 87.23 & \underline{0.8899}   & 7.99E-05    & 221.4786  \\
		& MWCNN \cite{MWCNN}    & 76.26 & 89.55    & 84.30  & 0.8685   & 0.0006      & 285.2711 \\
		& DUNet \cite{DUNet}   & 79.68 & 89.44    & 88.05 & 0.8874   & \underline{7.75E-05}    & 213.6870   \\
		& CAPS     & \textbf{81.66} & \underline{91.02}    & \textbf{89.94} & \textbf{0.9048}   & \textbf{1.62E-05}    & \textbf{46.4671}   \\     \hline
		\multirow{8}{*}{hazelnut}      & Baseline \cite{unet}& 80.47 & 93.77    & 85.92 & 0.8967   & 0.0005      & 128.2595  \\
		& BlurPool \cite{BlurPool}& \underline{82.86} & \underline{97.13}    & 85.06 & 0.9070   & 7.37E-05    & 75.8136   \\
		& APS \cite{APS}     & 80.28 & 93.78    & 84.84 & 0.8909   & 0.0008      & 98.6033   \\
		& LPS \cite{LPS}      & 77.19 & 90.86    & 84.82 & 0.8774   & 0.0027      & \underline{41.6838}   \\
		& PBP \cite{PBP}      & 81.24 & 95.65    & 84.88 & 0.8994   & 0.0004      & 209.5076  \\
		& MWCNN \cite{MWCNN}   & 80.91 & 94.74    & 85.00   & 0.8961   & 0.0005      & 112.8987  \\
		& DUNet \cite{DUNet}   & \textbf{83.76} & 96.41    & \underline{86.50}  & \underline{0.9119}   & \underline{5.76E-05}    & 50.3621   \\
		& CAPS     & 82.53 & \textbf{97.95}    & \textbf{86.63} & \textbf{0.9194}   & \textbf{4.32E-05} & \textbf{12.6327}   \\   \hline
		\multirow{8}{*}{photovoltaic modules} & Baseline \cite{unet}& 95.83 & 97.39    & 98.35 & 0.9787   & 0.0006      & 74.18678  \\
		& BlurPool \cite{BlurPool}& 96.86 & \textbf{98.02}    & 98.75 & 0.9838   & 0.0004      & 91.7825   \\
		& APS \cite{APS}     & 90.99 & 93.85    & 96.89 & 0.9535   & 0.0032      & \underline{52.3809}   \\
		& LPS \cite{LPS}      & 90.90 & 93.11    & 97.53 & 0.9527   & 0.0029      & 58.5660    \\
		& PBP \cite{PBP}      & 96.44 & 97.56    & 98.77 & 0.9816   & 0.0005      & 70.0917   \\
		& MWCNN \cite{MWCNN}   & 94.05 & 97.15    & 96.65 & 0.9690   & 0.0005      & 338.1548  \\
		& DUNet \cite{DUNet}    & \underline{97.25} & 97.73    & \textbf{99.52} & \textbf{0.9862}   & \underline{0.0003}      & 62.4016   \\
		& CAPS     & \textbf{97.27} & \underline{97.86}    & \underline{99.34} & \underline{0.9859}   & \textbf{0.0001}      & \textbf{42.2945}  \\
		\bottomrule 
	\end{tabular}
\begin{tablenotes}
	\item * The best results are shown in \textbf{bold} and the second best results are \underline{underlined}.
\end{tablenotes}
\end{threeparttable}
\end{table*}
Segmentation performance and shift equivalence for four datasets are quantitatively demonstrated in Table \ref{tab8}. The proposed CAPS achieves the best shift equivalence and remarkable segmentation performance compared with other methods. For shift equivalence, CAPS has the lowest mvIoU and mvda on all datasets, implying that the proposed method not only has the smallest IoU fluctuations, but also the highest stability of the predicted defect area. Moreover, it has the highest mIoU and f1-scores in 3 out of 4 datasets, showing its powerful defect segmentation ability. It can be observed that DUNet exhibits the best recall and f1-score on photovoltaic modules, and the second best recall and f1-score on hazelnut. But it slows down the inference speed as shown in Table \ref{tab6}, which is not suitable for industrial scenarios.

\section{CONCLUSION}
This paper presents a novel approach fousing on investigating shift equivalence of CNNs in industrial defect segmentation. The proposed method designs a pair of down/upsampling layers named CAPS to replace conventional downsampling and upsampling layers. The downsampling layer CAPD performs an attention-based fusion of the different components considering the feature boundaries. The CAPU then upsamples the downsampling results to a specific spatial location, ensuring the equivalence of the segmentation results. On the industrial defect segmentation test sets MDT and BDT, the proposed method surpasses other advanced methods such as BlurPool, APS, LPS, PBP, MWCNN and DUNet in terms of shift equivalence and segmentation performance.
\par

\bibliographystyle{IEEEtran}
\bibliography{IEEEabrv,reference.bib}

\end{document}